\documentclass[conference]{IEEEtran}
\IEEEoverridecommandlockouts
\usepackage{cite}
\usepackage{amsmath,amssymb,amsfonts}
\usepackage{algorithmic}
\usepackage{graphicx}
\usepackage{textcomp}
\usepackage{xcolor}
\usepackage{comment}
\usepackage{booktabs}
\usepackage{caption}
\usepackage{pgfplots}
\usepgfplotslibrary{groupplots}
\pgfplotsset{compat=1.18}
\def\BibTeX{{\rm B\kern-.05em{\sc i\kern-.025em b}\kern-.08em
    T\kern-.1667em\lower.7ex\hbox{E}\kern-.125emX}}
\begin{document}

\title{Diverse by Reasoning: Harnessing the Wisdom of LLM Crowds for Future Prediction
\thanks{*The first three authors have equal contributions}
}

\author{\IEEEauthorblockN{Nirupam Chetlapalli$^*$,Yiming Liao$^*$, Min-Chun Chen$^*$,Keke Chen}
\IEEEauthorblockA{\textit{Computer Science and Electrical Engineering} \\
\textit{University of Maryland, Baltimore County}\\
Baltimore, MD, USA \\
\{nchetla1, yliao1, mchen12, kekechen\}@umbc.edu}
}

\maketitle

\begin{abstract}
Large language models (LLMs) are increasingly used for future
prediction, motivating the use of multiple models as a
wisdom-of-the-crowd mechanism. However, simply increasing crowd size does not guarantee effective diversity, as different LLMs may exhibit redundant behaviors. We propose a behavior-aware framework for constructing diverse LLM crowds. The framework characterizes models using their reasoning traces on independent development tasks, clusters models by behavioral similarity, and selects representatives for collective prediction. We evaluate 25 LLMs using seven development benchmarks for behavioral diversity modeling and two future-prediction benchmarks for evaluating diverse crowds' performance. Our results show that crowd composition can matter more than crowd size: a three-model medoid
crowd based on $K$-means++ behavioral clustering outperforms
conventional voting over all 25 models on both prediction benchmarks, while reducing model calls by 88\% and inference cost by approximately 80\%. The results further suggest that representative behavioral diversity, rather than simply maximizing diversity, is important for constructing effective LLM crowds\footnote{The source code and datasets will be released.}.
\end{abstract}

\section{Introduction}
\label{sec:introduction}

Large language models (LLMs) are increasingly being explored for
forecasting uncertain future events~\cite{zou2022autocast,
halawi2024approaching}, with recent benchmarks providing systematic
evaluation of their forecasting capabilities
~\cite{karger2025forecastbench,wildman2025bench,zeng2026futurex}.
Despite this progress, future prediction remains intrinsically
uncertain, and individual LLMs can vary substantially in their
predictions. A natural alternative is therefore to aggregate multiple
LLMs as a \emph{crowd} of predictors.

The classical wisdom-of-the-crowd literature suggests that collective
judgment can outperform individual decision makers, but also emphasizes
that crowd composition---particularly diversity and independence---is
critical~\cite{surowiecki2004wisdom,goldstein2014smaller,
davisstober2015composition}. Carefully constructed smaller crowds can
even outperform much larger ones~\cite{galesic2018smaller,
bhatt2019captain}. Recent studies similarly demonstrate the potential
of collective intelligence among LLMs
~\cite{schoenegger2024silicon,chuang2025guesstimation}. However, simply
assembling different LLMs does not guarantee a genuinely diverse crowd.
Models exhibiting similar behaviors may contribute redundant judgments,
yet conventional majority voting gives each model an equal vote.

This motivates a complementary problem to existing work on LLM
aggregation~\cite{zhao2024electoral,ai2025beyond}: \emph{which models
should constitute the crowd in the first place?} A large crowd containing
many behaviorally similar models may over-represent particular
perspectives while incurring unnecessary inference cost. Conversely, a
smaller crowd that preserves distinct behavioral perspectives may yield
more effective collective predictions. We therefore ask:

\begin{quote}
\emph{Can we construct a wiser LLM crowd by explicitly accounting for
behavioral diversity rather than simply increasing the number of
voters?}
\end{quote}

We address this question through a behavior-driven approach. Rather than
inferring diversity from model family, provider, or other metadata, we
observe how LLMs reason over a common set of heterogeneous development
tasks. We embed their reasoning traces and aggregate them into
model-level \emph{behavioral signatures}, which are then clustered to
identify groups of models exhibiting similar behavior. Importantly,
these signatures are constructed entirely from development tasks
independent of the future-prediction benchmarks. We use the discovered
structure to construct crowds through representative medoids,
diversity-oriented selection, and cluster-level voting. This design
allows us to examine whether effective crowd construction requires
maximal behavioral diversity or, instead, representative coverage of
distinct behavioral modes.

We evaluate 25 LLMs using 350 questions from seven heterogeneous
development benchmarks and two independent future-prediction benchmarks,
FutureX-Past~\cite{zeng2026futurex} and Bench to the Future
~\cite{wildman2025bench}. Our results show that crowd composition can
matter more than crowd size. A three-model crowd constructed from
medoid representatives of $K$-means++ behavioral clusters outperforms
conventional voting over all 25 models on both prediction benchmarks,
while reducing model calls by 88\% and measured inference cost by
approximately 80\%. Size-matched random-crowd experiments further
support the benefit of behavior-aware selection. In contrast, maximizing
behavioral dissimilarity does not consistently improve prediction,
suggesting that \emph{representative behavioral diversity}, rather than
diversity alone, is important for constructing effective LLM crowds.

The main contributions of this work are:
\begin{itemize}
    \item We formulate LLM future prediction as a crowd-construction
    problem and identify behavioral redundancy as a limitation of
    conventional majority voting.

    \item We develop a behavior-aware framework that characterizes LLM
    diversity from reasoning traces on independent tasks and uses the
    resulting structure to construct representative crowds.

    \item Across 25 LLMs and two future-prediction benchmarks, we show
    that a three-model behavior-aware crowd outperforms conventional
    25-model voting while reducing model calls by 88\% and inference
    cost by approximately 80\%.

\end{itemize}
\section{Related Work}
\label{sec:related}

\textbf{Wisdom of the Crowd.}
The wisdom-of-the-crowd literature shows that aggregating multiple
judgments can outperform individual decision makers when the crowd
provides sufficiently diverse and independent information
~\cite{surowiecki2004wisdom}. Importantly, larger crowds are not
necessarily better: crowd composition can be optimized, and carefully
selected smaller crowds can match or outperform substantially larger
ones~\cite{goldstein2014smaller,davisstober2015composition,
galesic2018smaller}. This motivates treating crowd composition, rather
than crowd size alone, as a design problem.

Particularly relevant to our work, Bhatt et
al.~\cite{bhatt2019captain} inferred participant diversity from
observable behavior, clustered participants accordingly, and selected
representatives to construct smaller crowds. Our work extends this
behavior-driven view to LLMs: instead of using participant-generated
social content, we characterize models through their reasoning behavior
on heterogeneous tasks and examine whether the resulting structure can
guide crowd construction for future prediction.

\textbf{Collective Decision-Making with LLMs.}
Recent work has demonstrated several forms of collective intelligence
among LLMs. Self-consistency aggregates multiple reasoning trajectories
from a single model~\cite{wang2023selfconsistency}, while heterogeneous
LLM ensembles can improve forecasting and estimation through aggregation
across models~\cite{schoenegger2024silicon,chuang2025guesstimation}.
Other approaches introduce interaction among models, such as multi-agent
debate~\cite{du2024debate} and Mixture-of-Agents~\cite{wang2025moa}.
Recent studies have also reconsidered how heterogeneous model judgments
should be combined, including alternative electoral mechanisms
~\cite{zhao2024electoral} and aggregation rules that account for
higher-order information~\cite{ai2025beyond}.

These studies primarily improve how multiple model outputs are sampled,
combined, or exchanged. Our focus is complementary: we study
\emph{crowd construction}. Rather than assuming that different model
identities constitute a diverse crowd, we measure behavioral diversity
from reasoning traces and use the resulting structure to determine
which models should represent the crowd.

\textbf{LLM Future-Prediction Benchmarks.}
LLM forecasting has progressed from early benchmarks such as
Autocast~\cite{zou2022autocast} to systems approaching competitive human
forecasting performance~\cite{halawi2024approaching}. More recent
benchmarks address temporal contamination and realistic future
prediction through different evaluation designs. ForecastBench
maintains dynamically generated unresolved forecasting questions
~\cite{karger2025forecastbench}, Bench to the Future uses pastcasting
with period-correct information~\cite{wildman2025bench}, and FutureX
provides a continuously updated benchmark for evaluating LLM and agent
future prediction~\cite{zeng2026futurex}.

These benchmarks establish future prediction as a measurable LLM
capability, but primarily evaluate individual systems. Our work instead
uses future-prediction benchmarks to study a different question:
whether behavioral information obtained independently of the prediction
tasks can be used to construct a smaller and more effective crowd of
LLM predictors.

\section{Approach}
\label{sec:approach}

Our goal is to construct an LLM crowd that captures distinct behavioral
perspectives rather than simply including as many models as possible.
Conventional majority voting treats models as equally informative
voters, although their predictions and problem-solving behaviors may be
highly redundant. This can cause a large group of behaviorally similar
models to dominate the collective decision while contributing limited
additional information.

We therefore construct a \emph{behavior-aware diverse LLM crowd}. The key idea
is to infer diversity from how models behave on a common set of
development tasks, rather than from metadata such as model family,
provider, or parameter count. This follows the broader principle of
inferring crowd diversity from observable behavior
~\cite{bhatt2019captain}, but applies it to LLM reasoning behavior across
controlled tasks. The resulting behavioral structure is then used to
select representative voters or organize models into subcommittees.

The framework consists of three stages: (1) behavioral representation,
(2) behavioral clustering and crowd construction, and (3) prediction
aggregation. Importantly, the development questions used to characterize
behavior are separate from the future-prediction questions used for
evaluation. Thus, the crowd is constructed without using the labels of
the evaluation questions.

\subsection{Reasoning-Based Behavioral Representation}
\label{sec:behavior_representation}

A central design question is how to define diversity among LLMs.
Model identity provides a convenient proxy, but models from different
families may behave similarly, while models within the same family may
exhibit different problem-solving patterns. We therefore characterize
diversity directly from observable model behavior.

Given a common set of $T$ development questions, each model $M_j$
generates a reasoning trace $r_{ij}$ for question $q_i$. A text encoder
$E(\cdot)$ maps each trace to a vector, which is normalized to remove
differences in embedding magnitude. We summarize model $M_j$ by the
normalized average of its reasoning embeddings:

\begin{equation}
\mathbf{b}_j =
\operatorname{Norm}\left(
  \frac{1}{T}\sum_{i=1}^{T}
  \operatorname{Norm}\big(E(r_{ij})\big)
\right),
\label{eq:behavior_signature}
\end{equation}

where $\operatorname{Norm}(\mathbf{x})=\mathbf{x}/\|\mathbf{x}\|_2$.
We refer to $\mathbf{b}_j$ as the model's \emph{behavioral signature}.
As the vectors are derived from text data, where cosine similarity is often a better similarity measure, length normalization is used to facilitate cosine similarity calculation 
between these signatures.

This representation is intentionally simple. Averaging over many
heterogeneous questions emphasizes behavioral patterns that persist
across tasks rather than idiosyncratic responses to individual
questions. Normalization further prevents long or high-magnitude
embeddings from disproportionately affecting the signature. At the same
time, the signature should be interpreted as a representation of
\emph{observable reasoning behavior}, not as a faithful description of
a model's internal reasoning process.

This design also differs from approaches that characterize an LLM crowd
only through its final predictions. For example, recent aggregation
methods explicitly account for heterogeneity or correlations among model
answers~\cite{ai2025beyond}. Our objective is instead to obtain a
behavioral signal \emph{before} the downstream prediction task and test
whether this signal can be used to construct the crowd itself.

\subsection{Discovering Behavioral Structure}
\label{sec:behavior_clustering}

We use the behavioral signatures to identify groups of models exhibiting
similar reasoning behavior. Clustering serves an operational purpose:
it approximates different behavioral modes in the model population so
that a large group of similar models need not receive proportionally
greater representation in the final crowd. We do not interpret the
clusters as intrinsic or universal ``types'' of LLMs.

We determine the candidate number of clusters $K$ using the silhouette
coefficient \cite{Rousseeuw87} and examine two complementary clustering methods:
$K$-means++ and agglomerative hierarchical clustering. $K$-means++ provides
a centroid-based partition, while hierarchical clustering does not
depend on centroid initialization and provides a complementary view of
the behavioral structure. Using both methods also allows us to evaluate
whether downstream crowd performance depends strongly on a particular
clustering assumption.

This clustering stage distinguishes our approach from recent work that
introduces diversity primarily through the collective decision rule.
For example, Zhao et al.~\cite{zhao2024electoral} diversify multi-agent
decision making through alternative electoral mechanisms. In contrast,
we retain simple aggregation rules and modify the \emph{composition of
the crowd} based on behavior observed independently of the prediction
task.

\subsection{Behavior-Aware Crowd Construction}
\label{sec:crowd_construction}

Given behavioral clusters $C_1,\ldots,C_K$, we consider three strategies
that represent different interpretations of useful crowd diversity.

\subsubsection{Medoid Representatives}

The first strategy selects one representative from each behavioral
cluster. For cluster $C_c$, we choose its medoid,

\begin{equation}
M_c^{\mathrm{med}}
=
\arg\min_{M_i\in C_c}
\sum_{M_j\in C_c} d(M_i,M_j),
\label{eq:medoid}
\end{equation}

where $d(\cdot,\cdot)$ is the behavioral distance derived from the
signatures.

The medoid is an actual LLM whose observed behavior is most
representative of its cluster. Selecting one medoid per cluster has two
motivations. First, it preserves coverage of the discovered behavioral
groups while reducing the crowd from $N$ models to $K$. Second, each
cluster receives one vote regardless of how many behaviorally similar
models it contains, reducing the effect of behavioral redundancy.

This strategy is related in spirit to behavior-based crowd construction
in human crowds~\cite{bhatt2019captain}, but differs in both the
behavioral signal and selection objective. Here, diversity is inferred
from LLM reasoning traces over controlled benchmark tasks, and the
medoid strategy emphasizes \emph{representative coverage} rather than
maximal diversity.

\subsubsection{Diverse Representatives}

A medoid captures the central behavior of a cluster but may discard
meaningful within-cluster variation. We therefore consider a second
strategy that selects $k$ representatives from each cluster to maximize
their pairwise behavioral distances:

\begin{equation}
R_c^{*}
=
\arg\max_{\substack{R\subseteq C_c\\|R|=k}}
\sum_{\substack{M_i,M_j\in R\\i<j}} d(M_i,M_j).
\label{eq:diverse_representatives}
\end{equation}

This strategy provides an important contrast to medoid selection.
Whereas medoids seek \emph{representativeness}, the $k$-dissimilar
strategy explicitly favors \emph{diversity}. Comparing the two allows
us to test whether crowd wisdom benefits from maximizing behavioral
differences or from ensuring representative coverage of distinct
behavioral groups. We use small values of $k$; the specific settings
are described in Section~\ref{sec:experiments}.

\subsubsection{Cluster Subcommittees}

The third strategy retains all models but changes how their votes are
counted. Each behavioral cluster acts as a local subcommittee:

\begin{equation}
\hat{y}_c(q)
=
\operatorname{Vote}
\{a_j(q):M_j\in C_c\}.
\label{eq:cluster_vote}
\end{equation}

The resulting cluster decisions are then aggregated globally. Thus, a
cluster containing many similar models does not automatically receive
more influence than a smaller behavioral cluster.

Unlike medoid and diverse-representative selection, this strategy does
not reduce the number of model calls. Instead, it isolates the effect
of \emph{vote balancing}: if behavioral clusters approximate distinct
sources of information, treating each cluster as one higher-level voter
may prevent redundant models from dominating the result. This provides
a direct test of whether behavioral groups, rather than individual
models, should be treated as the effective units of crowd diversity.

\subsection{Prediction Aggregation}
\label{sec:aggregation}

For the \textbf{medoid} and \textbf{diverse-representative} strategies, selected models
vote as follows:

\begin{equation}
\hat{y}(q)
=
\operatorname{Vote}
\{a_j(q):M_j\in\mathcal{R}\},
\label{eq:representative_vote}
\end{equation}
where $\mathcal{R}$ is the selected representative set. 

The actual aggregation method used in $\operatorname{Vote}$ can be different according to the answer types. The benchmarks in general may contain four heterogeneous answer types. We use a type-aware aggregation procedure rather than applying a single voting rule to all questions. (1) For categorical answers, including binary and (N)-choose-one questions, we use majority (plurality) voting and select the answer receiving the most votes. BTF-v3 and Level-1 questions for FutureX-past are in this category. (2) For numeric answers, we use the median, which provides a robust consensus estimate without being overly influenced by extreme predictions. (3) For set-valued answers, e.g., multiple-choice questions, where exact agreement may be unnecessarily restrictive, we use similarity-based medoid aggregation: pairwise agreement between two answers is measured by set (i.e., the $F_1$ measure for answers $A_i$ and $A_j$: $F_1(A_i,A_j)
=
\frac{2|A_i \cap A_j|}
{|A_i| + |A_j|}$ ), and then we select the observed answer with the highest average similarity to all other crowd answers (i.e., the medoid principle). The Level-2 questions in FutureX-past are in this category. (4) Finally, for ordered-list answers, e.g., the Level-3/4 questions in FutureX-past, we apply the same medoid principle for all pairs of ordered-list answers, and use the OrderedOverlap pairwise similarity, which accounts for agreement in both the selected elements and their ordering. We adopted the OrderedOverlap similarity function provided by the FutureX-past benchmark. More generally, for the latter two structured-output cases, given an answer-specific similarity function $S(A_i,A_j)$, where $A_i$ and $A_j$ are two answers, the aggregate is selected as $\hat{A}=\arg\max_{A_i}\frac{1}{N-1}\sum_{j\ne i}S(A_i,A_j)$. This type-aware design preserves the natural structure of each answer space while providing a common consensus principle for outputs for which exact-match voting is inappropriate.

For \textbf{cluster subcommittees}, the same rule is applied hierarchically:
models first vote within each cluster according to
Eq.~\eqref{eq:cluster_vote}, and the final prediction is obtained by
voting over the $K$ cluster decisions.

We deliberately use simple aggregation rules so that improvements can
be attributed primarily to \emph{crowd construction} rather than to a
more sophisticated voting mechanism. This separates our question from
work that improves collective decisions through electoral rules
~\cite{zhao2024electoral} or higher-order aggregation
~\cite{ai2025beyond}. It also allows a direct comparison with standard
all-model majority voting, which has been shown to provide useful
collective forecasting performance for heterogeneous LLMs
~\cite{schoenegger2024silicon}.

Overall, the three strategies test different hypotheses about the source
of crowd wisdom: medoid selection emphasizes representative behavioral coverage,
dissimilar selection emphasizes maximal behavioral diversity, and
subcommittee voting emphasizes balanced influence across behavioral
groups. Their empirical comparison therefore allows us to examine not
only whether behavioral diversity matters, but \emph{what form of
diversity is most useful for LLM crowds}.

\section{Experiments}
\label{sec:experiments}

Our experiments investigate whether behavioral diversity observed on general reasoning tasks can be used to construct effective crowds for future prediction. A key aspect of our experimental design is the separation between \emph{behavioral characterization} and
\emph{future-prediction evaluation}. We use seven general-purpose benchmarks solely to characterize the reasoning behavior of the LLM population and construct behaviorally diverse crowds. The resulting crowds are then evaluated on two separate future-prediction benchmarks.
No future-prediction question is used to construct the behavioral representations or clusters.

We organize the evaluation around three research questions:
\begin{itemize}
    \item \textbf{RQ1---Behavioral Diversity:}
    Can reasoning traces reveal meaningful and stable behavioral
    diversity among LLMs?

    \item \textbf{RQ2---Prediction Effectiveness:}
    Can behaviorally diverse crowds improve future-prediction
    performance?

    \item \textbf{RQ3---Prediction Efficiency:}
    Can behavioral diversity preserve prediction quality with
    substantially fewer model calls?
\end{itemize}

Together, these questions examine whether behavioral diversity exists, whether it is useful, and whether it can be exploited to construct a more efficient crowd.

\subsection{Experimental Setup}
\label{sec:experimental_setup}

\subsubsection{Models}

Our model population consists of $N=25$ LLMs. The collection includes
models from different model families and providers in order to represent
a heterogeneous population of prediction agents. Table~\ref{tab:model_list} lists them
with their training cutoffs, which Section~\ref{sec:benchmarks} draws on.

\begin{itemize}
\item \textbf{Open-weight models (17)}:
Gemma-3-12B-IT, Gemma-3-27B-IT;
Llama-3.3-70B-Instruct, Llama-4-Maverick;
Qwen-2.5-72B-Instruct, Qwen3.5-397B-A17B;
DeepSeek-Chat-v3.1, DeepSeek-v3.2;
Kimi-K2.5, Kimi-K2.6;
GLM-4.7, GLM-5;
MiniMax-M2.5, MiniMax-M3;
Mistral-Large-2512;
Nemotron-3-Super-120B-A12B, Nemotron-3-Ultra-550B-A55B.

\item \textbf{Closed-weight models (8)}:
GPT-5.4, GPT-5.4-mini;
Claude-Haiku-4.5, Claude-Sonnet-4.6;
Gemini-3.1-Flash-Lite, Gemini-3.1-Pro-Preview;
Grok-4.3;
Mistral-Medium-3.1.
\end{itemize}


\begin{table*}[t]
\caption{The $N=25$ LLMs in the experimental crowd, grouped by developer.
\textbf{Cutoff / Release} is the training cutoff where the provider publishes one (16 of
the 25 models), and the release date otherwise (9). A model cannot be trained on data that
did not yet exist when it was released, so for those 9 the real cutoff can only be earlier
than the date shown. The latest date across the crowd is 2026--02--16.}
\label{tab:model_list}
\centering
\small
\begin{tabular}{@{}lll@{\hspace{2.5em}}lll@{}}
\toprule
\textbf{Model} & \textbf{Weights} & \textbf{Cutoff / Release} & \textbf{Model} & \textbf{Weights} & \textbf{Cutoff / Release} \\
\midrule
\multicolumn{3}{@{}l}{\itshape OpenAI} & \multicolumn{3}{@{}l}{\itshape DeepSeek} \\
\quad gpt-5.4 & Closed & 2025--08--31 & \quad deepseek-chat-v3.1 & Open & 2025--03--31 \\
\quad gpt-5.4-mini & Closed & 2025--08--31 & \quad deepseek-v3.2 & Open & 2025--12--01 \\
\multicolumn{3}{@{}l}{\itshape Anthropic} & \multicolumn{3}{@{}l}{\itshape Moonshot AI} \\
\quad claude-haiku-4.5 & Closed & 2025--10 & \quad kimi-k2.5 & Open & 2026--01--27 \\
\quad claude-sonnet-4.6 & Closed & 2026--01 & \quad kimi-k2.6 & Open & 2025--04 \\
\multicolumn{3}{@{}l}{\itshape Google} & \multicolumn{3}{@{}l}{\itshape Z.AI} \\
\quad gemini-3.1-flash-lite & Closed & 2025--01 & \quad glm-4.7 & Open & 2025--07 \\
\quad gemini-3.1-pro-preview & Closed & 2025--01 & \quad glm-5 & Open & 2026--01 \\
\quad gemma-3-12b-it & Open & 2024--08--31 & \multicolumn{3}{@{}l}{\itshape MiniMax} \\
\quad gemma-3-27b-it & Open & 2024--08--31 & \quad minimax-m2.5 & Open & 2025--01 \\
\multicolumn{3}{@{}l}{\itshape xAI} & \quad minimax-m3 & Open & 2026--01 \\
\quad grok-4.3 & Closed & 2025--11 & \multicolumn{3}{@{}l}{\itshape Mistral AI} \\
\multicolumn{3}{@{}l}{\itshape Meta} & \quad mistral-large-2512 & Open & 2025--12--01 \\
\quad llama-3.3-70b-instruct & Open & 2023--12--31 & \quad mistral-medium-3.1 & Closed & 2025--06--30 \\
\quad llama-4-maverick & Open & 2024--08--31 & \multicolumn{3}{@{}l}{\itshape NVIDIA} \\
\multicolumn{3}{@{}l}{\itshape Alibaba} & \quad nemotron-3-super-120b-a12b & Open & 2025--12 \\
\quad qwen-2.5-72b-instruct & Open & 2024--06--30 & \quad nemotron-3-ultra-550b-a55b & Open & 2025--12 \\
\quad qwen3.5-397b-a17b & Open & 2026--02--16 &  &  &  \\
\bottomrule
\end{tabular}
\end{table*}
For every question, each model is queried independently. We collect both its final answer and its generated reasoning trace. The latter is used only for constructing the behavioral representation described in
Section~\ref{sec:behavior_representation}.


\subsubsection{Development Benchmarks}
To characterize general reasoning behavior, we use seven development benchmarks spanning heterogeneous capabilities\footnote{LiveBench datasets: https://huggingface.co/livebench (updated on Apr 7, 2025); GPQA: epoch.ai/benchmarks/use-this-data; Natural Plan: https://github.com/google-deepmind/natural-plan, and LiveCodeBench: livecodebench.github.io }: LiveBench/Reasoning, LiveBench/Math, LiveBench/Instruction-Following,
    LiveBench/Data-Analysis,
    GPQA Diamond,
    Natural Plan, and
    LiveCodeBench/Execution-v2.

We randomly sample 50 questions from each benchmark, producing a total of 350 development questions. Using the same number of questions from each benchmark prevents any single benchmark from dominating the behavioral
representation simply because of its size.

The seven benchmarks intentionally cover different reasoning capabilities, including general and mathematical reasoning, scientific reasoning, instruction following, data analysis, planning, and program
execution. Our purpose is not to evaluate model capability on these benchmarks per se. Instead, they serve as a heterogeneous set of \emph{behavioral probes} for observing systematic differences among LLMs.

All 25 models receive the same set of development questions. Their reasoning traces on these questions are embedded and aggregated to construct the model-level behavioral signatures.

\subsubsection{Future-Prediction Benchmarks}
\label{sec:benchmarks}
We evaluate the constructed crowds on two future-prediction benchmarks\footnote{huggingface.co/datasets/futurex-ai/Futurex-Past and huggingface.co/datasets/BTF-2/BTF-3}: FutureX-Past and Bench to the Future v3 (BTF-v3). \paragraph{Temporal separation and training-data leakage}
Both benchmarks consist of questions about events that had not yet occurred when they were
written, and both are scored against outcomes that are now known. This makes the temporal
relationship between a model's training data and a question's resolution decisive rather than
incidental: a model whose training corpus already contains the outcome is not forecasting it,
it is recalling it, and any score it earns that way measures memorisation instead of
prediction. The two are indistinguishable from the answer alone, so the separation has to be
established by construction.

The latest training cutoff among the 25 models is
16~February 2026 (Table~\ref{tab:model_list}), and we require every evaluation question to
resolve after that date. In our samples the FutureX-Past questions resolve
between 7~March and 28~July 2026, and the BTF-v3 questions between 5~May and 1~July 2026;
both windows open after that date, giving a margin of at least three weeks for FutureX-Past
and ten weeks for BTF-v3. Candidate questions resolving on or before it were removed from the
pool before sampling rather than filtered afterwards, so the guarantee does not depend on
the sample that happened to be drawn. No question in either benchmark can have been observed
during any model's training.

We randomly sample 100 questions from each benchmark, resulting in 200 future-prediction questions. These questions are completely separated from the development benchmarks: they are not used to construct behavioral signatures, determine the number of clusters, or select
cluster representatives.

For each benchmark, we adopt the question prompts specified by the original benchmark authors without introducing additional prompt engineering. This controls for prompt-related factors and allows the experiments to focus on the effect of crowd construction and
aggregation.

Table~\ref{tab:benchmark_summary} summarizes the datasets used in the
study.

\begin{table*}[t]
\centering
\caption{Benchmarks used for behavioral characterization (top 7 rows) and
future-prediction evaluation. \textbf{Resolves} is the range of dates on which the questions'
outcomes became known, for direct comparison with the training cutoffs in
Table~\ref{tab:model_list}: the earliest of these, 2026--03--07, falls after the latest cutoff in
the crowd, 2026--02--16. The development benchmarks are not time-indexed and have no resolution
date.}
\label{tab:benchmark_summary}
\small
\begin{tabular}{llll}
\toprule
\textbf{Benchmark} &
\textbf{\# Questions} &
\textbf{Capability} &
\textbf{Resolves} \\
\midrule
LiveBench/Reasoning & 50 & General reasoning & --- \\
LiveBench/Math & 50 & Mathematical reasoning & --- \\
 LiveBench/Instruction-Following & 50 & Instruction following & --- \\
LiveBench/Data-Analysis & 50 & Data analysis & --- \\
GPQA Diamond & 50 & Scientific reasoning & --- \\
Natural Plan & 50 & Planning & --- \\
 LiveCodeBench/Execution-v2 & 50 & Program execution & --- \\
\midrule
FutureX-Past & 100 & Future prediction & 2026--03--07 to 2026--07--28 \\
Bench to the Future v3 (BTF-v3) & 100 & Future prediction & 2026--05--05 to 2026--07--01 \\
\bottomrule
\end{tabular}
\end{table*}

\subsubsection{Evaluation Metrics}
\label{sec:metrics}

The two benchmarks elicit different answer types and are scored by their own rules.

\textbf{Bench to the Future v3 (BTF-v3)} elicits a probability that a binary event occurs. We report accuracy
at a $0.5$ threshold: $p \geq 0.5$ ``yes'', otherwise, ``no''.

\textbf{FutureX-Past} has four difficulty levels of questions.  Level~1 is scored by exact match, i.e., ``yes''/``no'' answers. A level~2 answer contains a set, e.g., multiple-select answers. It is evaluated by computing the $F_1$ between the given answer and the correct answer.  Levels~3 and~4 give ordered list, e.g., a ranking. The evaluation awards $1.0$ for an exactly ordered match and $0.8\times|\text{overlap}|/|g|$
otherwise. Per-level means are combined with the benchmark's own weights of $10$, $20$, $30$
and $40$. We have completely adopted the FutureX's answer scoring function without modification. 

\textbf{Abstention and denominators.} Two ways of producing no answer are priced differently,
because they have different causes. A cell is \emph{unusable} when a model returned nothing
interpretable: this is outside the method's control, and following the FutureX-Past protocol such a
question is dropped from both sides of the score rather than counted as wrong. A crowd
\emph{abstains} when its members answered but reached no decision---at level~1 when the plurality
vote ties, and at any level when no member of the crowd responded at all. An abstention is the
method's own output and is kept in the denominator as a miss; excluding it would let a crowd raise
its score by declining exactly the questions its members disagree about, which are the hard ones.
At levels~2--4 the medoid rule always returns one of the observed answers, so a crowd never abstains
there; ties among equally central candidates are resolved as described later.

On FutureX-Past, $40$ of the $2{,}500$ model--question cells in the $25$-model roster are unusable:
$22$ empty, $15$ uninterpretable, and $3$ for which no response was returned at all. BTF-v3 has none.
The effect on the crowd methods is negligible---all-model voting loses no question, and no committee
loses more than one---so their denominators are the full $100$ questions in every case reported here.

\textbf{A common notation.} We write $s(\hat{y}, y)$ for the per-question score of a prediction
$\hat{y}$ against the resolved outcome $y$: the indicator $\mathbb{I}[\hat{y}=y]$ on Bench to the
Future v3, and the per-level score above on FutureX-Past. For a method $M$ over a question set
$\mathcal{Q}$ we write
\begin{equation}
S(M, \mathcal{Q}) = \operatorname{agg}_{q\in\mathcal{Q}} s\!\left(\hat{y}_M(q), y(q)\right),
\label{eq:score}
\end{equation}
where $\operatorname{agg}$ is the mean on BTF-v3 and the tier-weighted mean on
FutureX-Past. 

\subsubsection{Behavioral Representation and Clustering}
For each model $M_j$ and development question $q_i$, we encode the generated reasoning trace using \texttt{BAAI/bge-m3}. The default number of embedding dimensions is 1024. Each
question-level embedding is normalized to unit length to avoid overweighing questions. Following Section~\ref{sec:behavior_representation}, we obtain the behavioral
signature of model $M_j$ by averaging its normalized embeddings across
all 350 development questions and normalizing the resulting model-level
vector.

We evaluate candidate cluster counts $K \in \{2,\ldots, 8 \}$ and use the silhouette score to identify the clustering structure most
strongly supported by the behavioral representations.

We apply both $K$-means++ with 50 restarts and agglomerative
hierarchical clustering. Unless otherwise specified, the
silhouette-selected value of $K$ is used for downstream crowd construction.

\subsubsection{Compared Methods}

We compare the following prediction strategies.


\textbf{All-Model Voting.}
All 25 models participate in majority voting with equal weight. This is
our primary conventional wisdom-of-the-crowd baseline.

\textbf{Performance-Based Expert.}
A single expert is selected as the highest-scoring model on the selection folds, and then
evaluated on the held-out fold. The future-prediction labels are not used for expert
selection.

\textbf{Random Representatives.}
We randomly select the same number of models as used by the
corresponding behavior-aware representative method and aggregate their
predictions through majority voting. Random selection is repeated
$1000$ times.

\textbf{Medoid Representatives.}
One medoid is selected from each behavioral cluster, producing $K$ representatives. Their predictions are combined
through majority voting.

\textbf{$k$-Dissimilar Representatives.}
We select $k$ behaviorally dissimilar representatives from each cluster and perform majority voting over the resulting crowd. The purpose is to maximize the within-cluster diversity, while the overall number of representatives is kept to less than half of the total population. Thus, for $K=3$, we choose 
$k=4$.

\textbf{Cluster Subcommittees.}
All models participate, but voting is hierarchical. Models first vote
within their behavioral cluster, after which the $K$ cluster decisions
participate equally in the global vote.

\subsubsection{Evaluation Protocol}
\label{sec:evaluation_protocol}
We evaluate all prediction methods separately on FutureX-Past and BTF-v3. Each
benchmark contains 100 sampled future-prediction questions.

To provide a common evaluation protocol and to prevent any label-dependent selection procedure from
using the questions on which it is evaluated, we employ five-fold cross-validation independently on
each future-prediction benchmark. The 100 questions are partitioned into five folds of exactly 20 by
a seeded shuffle (seed $20260819$); questions are sorted before shuffling so that the order in which
responses were collected cannot influence the split. For fold $f$, the remaining four folds (80
questions) form the selection set $\mathcal{Q}_{\mathrm{sel}}^{(f)}$, while fold $f$ serves as the
held-out evaluation set $\mathcal{Q}_{\mathrm{test}}^{(f)}$. The same split is used for every method
and both benchmarks, so all comparisons are paired on identical questions.

In particular, for the performance-based expert baseline, the expert for fold $f$ is selected as
\begin{equation}
M_f^{*}
=
\arg\max_{M_j\in\mathcal{M}}
S\!\left(M_j,\ \mathcal{Q}_{\mathrm{sel}}^{(f)}\right),
\label{eq:expert_selection}
\end{equation}
with $S$ as defined in Eq.~\eqref{eq:score}, and $M_f^{*}$ is subsequently evaluated only on
$\mathcal{Q}_{\mathrm{test}}^{(f)}$. Ties on the selection set are broken by the finer-grained score of the same benchmark and then
by model name, so the choice is deterministic.

In contrast, our behavioral clusters are constructed exclusively from the 350 development-benchmark
questions and therefore remain independent of the future-prediction labels. Similarly, medoid and
diversity-based representatives are determined from these behavioral clusters and behavioral
distances rather than from the labels of the future-prediction questions. Consequently, these
components remain fixed across the five evaluation folds, and for those methods the pooled held-out
score is identical to the score they would obtain on all 100 questions at once. Cross-validation is
therefore not a safeguard for the behavior-aware methods but for the expert baseline they are
compared against, which is the only method whose construction consumes labels.

For each fold, all competing methods are evaluated on the same held-out questions. After completing
all five folds, we pool the held-out predictions so that every question contributes exactly once to
the reported evaluation. Thus the reported benchmark-level score is
\begin{equation}
S
=
\operatorname*{agg}_{f=1}^{5}\ \operatorname*{agg}_{q\in\mathcal{Q}_{\mathrm{test}}^{(f)}}
s\!\left(\hat{y}^{(f)}(q),\, y(q)\right),
\label{eq:cv_score}
\end{equation}
where $s(\cdot)$ and $\operatorname{agg}$ are as defined in Section~\ref{sec:metrics}. On BTF-v3 this reduces to the mean of 100 indicator values. On FutureX-Past the tier weights are
applied once, to the pooled 100 predictions, rather than per fold: a weighted mean of five
fold-level weighted means is not the weighted mean of the pooled set, and individual folds need not
contain every difficulty level.

This procedure yields 100 held-out predictions for each method on each future-prediction benchmark,
rather than averaging five scores computed from only 20 questions each. We used OpenRouter APIs for all the LLM calls. 


\subsection{RQ1: Behavioral Diversity among LLMs}
\label{sec:rq1}

Our first question examines whether the reasoning traces produced by different LLMs reveal meaningful structure in the model population. This is a prerequisite for the proposed approach: if the behavioral
representations do not exhibit systematic diversity, cluster-based
crowd construction would have little justification.

\subsubsection{Clustering Structure}
\label{sec:clustering_structure}
\begin{figure}[h]
\centering
\begin{tikzpicture}
\begin{axis}[
    width=0.75\linewidth,
    height=5.5cm,
    xlabel={Number of clusters ($K$)},
    ylabel={Silhouette score (SC)},
    xmin=2, xmax=8,
    ymin=0.18, ymax=0.35,
    xtick={2,3,4,5,6,7,8},
    ytick={0.18,0.22,0.26,0.30,0.34},
    grid=major,
    mark size=2.5pt,
]

\addplot[
    thick,
    mark=*,
]
coordinates {
    (2,0.334)
    (3,0.237)
    (4,0.196)
    (5,0.196)
    (6,0.213)
    (7,0.208)
    (8,0.215)
};

\addplot[
    only marks,
    mark=*,
    mark size=4pt,
]
coordinates {(2,0.334)};

\node[
    anchor=north west,
    font=\small
] at (axis cs:2,0.334) {$K=2$};

\addplot[
    only marks,
    mark=square*,
    mark size=4pt,
]
coordinates {(3,0.237)};

\node[
    anchor=south west,
    font=\small
] at (axis cs:3,0.237) {$K=3$};

\end{axis}
\end{tikzpicture}

\caption{Silhouette score of $K$-means++ across different numbers
of clusters. While $K=2$ has the highest SC, it may lead to many ties for medoid voting. Thus, $K=3$ is selected as a more practical choice.}
\label{fig:best-k}
\end{figure}
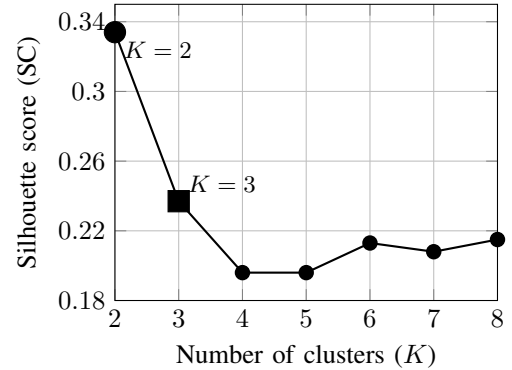

We sweep the number of clusters $K$ from 2 to 8 with $K$-means++ and select it using the silhouette score (SC) \cite{Rousseeuw87}. Figure \ref{fig:best-k} shows that SC reaches its peak at $K=2$, which, however, is not optimal for our purpose. Note that the medoid method yields only two representatives at $K=2$, and a similarity medoid over two candidates is tied by construction. Section~\ref{sec:choice_of_k} measures what this costs: at $K=2$ a tie decides 72 of the 100 FutureX-Past questions, against 28--32 at $K=3$. To avoid this, we consider the next-best structure with $K\geq 3$. Among the nontrivial solutions ($K \geq 3$), $K=3$ achieves the highest silhouette score (0.237) while retaining reasonably high clustering stability (0.767). We therefore select $K=3$ as a parsimonious solution that balances cluster separation, reproducibility, and the need for multiple representative perspectives.

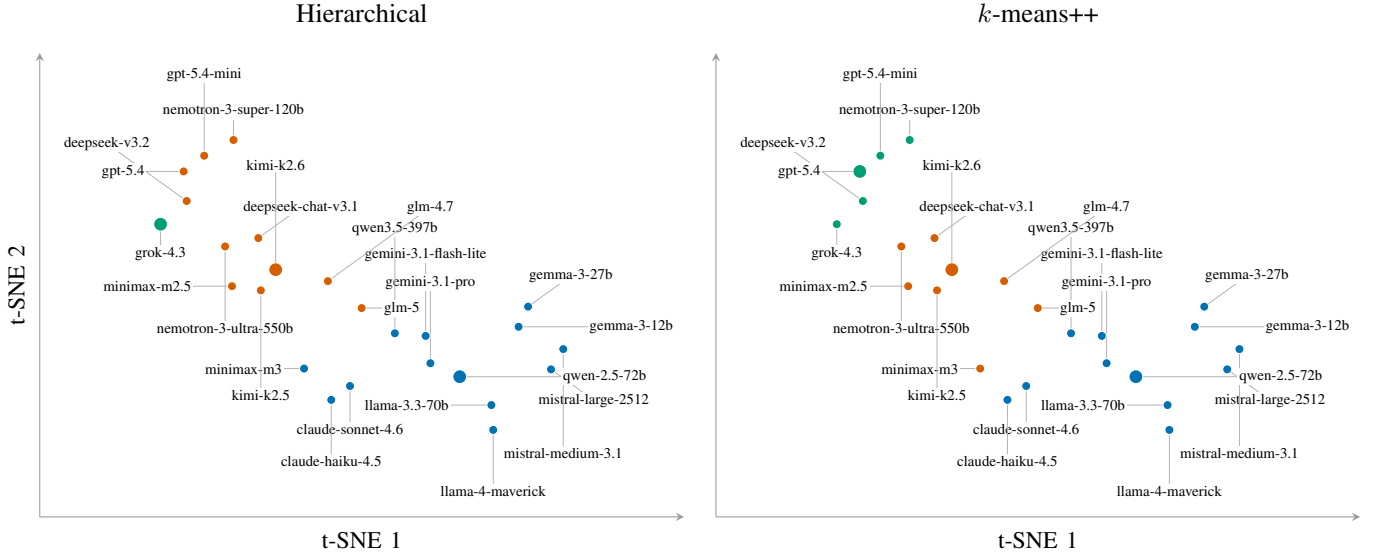
\begin{figure*}[t]
  \centering
  \begin{tikzpicture}
    \definecolor{cluster0}{HTML}{0072B2}
    \definecolor{cluster1}{HTML}{D55E00}
    \definecolor{cluster2}{HTML}{009E73}
    \begin{groupplot}[
      group style={group size=2 by 1, horizontal sep=12pt,
                    ylabels at=edge left, xlabels at=edge bottom},
      width=0.47\textwidth, height=0.3384\textwidth,
      axis lines=left, clip=false, scale only axis,
      axis line style={draw=black!40, line width=0.4pt},
      xlabel={t-SNE 1}, ylabel={t-SNE 2},
      label style={font=\small, text=black},
      xtick=\empty, ytick=\empty,
      enlargelimits=0.3, title style={font=\normalsize, text=black, yshift=2pt},
    ]
    \nextgroupplot[title={Hierarchical}]
      \addplot[only marks, mark=*, mark size=1.7pt, color=cluster0, mark options={fill=cluster0, draw=white, line width=0.4pt}] coordinates {(37.6480,-30.8920) (37.1715,-24.1102) (44.1936,-2.7490) (46.6350,2.7476) (52.5495,-14.3696) (55.7034,-8.8385) (-11.1189,-14.1490) (12.2934,-4.5243) (-4.1004,-22.6952) (0.7373,-18.9073) (20.2308,-5.2400) (21.4558,-12.6927)};
      \addplot[only marks, mark=*, mark size=2.6pt, color=cluster0, mark options={fill=cluster0, draw=white, line width=0.4pt}] coordinates {(29.0138,-16.3732)};
      \addplot[only marks, mark=*, mark size=1.7pt, color=cluster1, mark options={fill=cluster1, draw=white, line width=0.4pt}] coordinates {(-29.6915,8.3274) (-22.2548,7.1923) (-22.9018,21.4683) (-41.3660,31.5803) (-4.9664,9.7177) (3.7250,2.3623) (-29.2996,48.2286) (-31.4614,19.1747) (-42.1474,39.6493) (-36.8546,43.9386)};
      \addplot[only marks, mark=*, mark size=2.6pt, color=cluster1, mark options={fill=cluster1, draw=white, line width=0.4pt}] coordinates {(-18.4294,12.8358)};
      \addplot[only marks, mark=*, mark size=2.6pt, color=cluster2, mark options={fill=cluster2, draw=white, line width=0.4pt}] coordinates {(-48.0956,25.2472)};
      \draw[gray!60, line width=0.3pt] (axis cs:-29.2996,48.2286) -- (axis cs:-29.2996,56.1407);
      \node[font=\fontsize{5.6}{6.4}\selectfont, text=black, anchor=center, fill=white, inner sep=0.6pt] at (axis cs:-29.2996,56.1407) {nemotron-3-super-120b};
      \draw[gray!60, line width=0.3pt] (axis cs:-31.4614,19.1747) -- (axis cs:-31.4614,-2.9791);
      \node[font=\fontsize{5.6}{6.4}\selectfont, text=black, anchor=center, fill=white, inner sep=0.6pt] at (axis cs:-31.4614,-2.9791) {nemotron-3-ultra-550b};
      \draw[gray!60, line width=0.3pt] (axis cs:20.2308,-5.2400) -- (axis cs:20.2308,16.9138);
      \node[font=\fontsize{5.6}{6.4}\selectfont, text=black, anchor=center, fill=white, inner sep=0.6pt] at (axis cs:20.2308,16.9138) {gemini-3.1-flash-lite};
      \draw[gray!60, line width=0.3pt] (axis cs:52.5495,-14.3696) -- (axis cs:63.5590,-22.7616);
      \node[font=\fontsize{5.6}{6.4}\selectfont, text=black, anchor=center, fill=white, inner sep=0.6pt] at (axis cs:63.5590,-22.7616) {mistral-large-2512};
      \draw[gray!60, line width=0.3pt] (axis cs:55.7034,-8.8385) -- (axis cs:55.7034,-37.3219);
      \node[font=\fontsize{5.6}{6.4}\selectfont, text=black, anchor=center, fill=white, inner sep=0.6pt] at (axis cs:55.7034,-37.3219) {mistral-medium-3.1};
      \draw[gray!60, line width=0.3pt] (axis cs:-22.9018,21.4683) -- (axis cs:-11.8923,29.8603);
      \node[font=\fontsize{5.6}{6.4}\selectfont, text=black, anchor=center, fill=white, inner sep=0.6pt] at (axis cs:-11.8923,29.8603) {deepseek-chat-v3.1};
      \draw[gray!60, line width=0.3pt] (axis cs:0.7373,-18.9073) -- (axis cs:0.7373,-30.7754);
      \node[font=\fontsize{5.6}{6.4}\selectfont, text=black, anchor=center, fill=white, inner sep=0.6pt] at (axis cs:0.7373,-30.7754) {claude-sonnet-4.6};
      \draw[gray!60, line width=0.3pt] (axis cs:37.6480,-30.8920) -- (axis cs:37.6480,-47.5074);
      \node[font=\fontsize{5.6}{6.4}\selectfont, text=black, anchor=center, fill=white, inner sep=0.6pt] at (axis cs:37.6480,-47.5074) {llama-4-maverick};
      \draw[gray!60, line width=0.3pt] (axis cs:-4.1004,-22.6952) -- (axis cs:-4.1004,-39.3105);
      \node[font=\fontsize{5.6}{6.4}\selectfont, text=black, anchor=center, fill=white, inner sep=0.6pt] at (axis cs:-4.1004,-39.3105) {claude-haiku-4.5};
      \draw[gray!60, line width=0.3pt] (axis cs:21.4558,-12.6927) -- (axis cs:21.4558,9.4611);
      \node[font=\fontsize{5.6}{6.4}\selectfont, text=black, anchor=center, fill=white, inner sep=0.6pt] at (axis cs:21.4558,9.4611) {gemini-3.1-pro};
      \draw[gray!60, line width=0.3pt] (axis cs:37.1715,-24.1102) -- (axis cs:15.3737,-24.1102);
      \node[font=\fontsize{5.6}{6.4}\selectfont, text=black, anchor=center, fill=white, inner sep=0.6pt] at (axis cs:15.3737,-24.1102) {llama-3.3-70b};
      \draw[gray!60, line width=0.3pt] (axis cs:-41.3660,31.5803) -- (axis cs:-61.9172,47.2454);
      \node[font=\fontsize{5.6}{6.4}\selectfont, text=black, anchor=center, fill=white, inner sep=0.6pt] at (axis cs:-61.9172,47.2454) {deepseek-v3.2};
      \draw[gray!60, line width=0.3pt] (axis cs:-29.6915,8.3274) -- (axis cs:-51.4893,8.3274);
      \node[font=\fontsize{5.6}{6.4}\selectfont, text=black, anchor=center, fill=white, inner sep=0.6pt] at (axis cs:-51.4893,8.3274) {minimax-m2.5};
      \draw[gray!60, line width=0.3pt] (axis cs:29.0138,-16.3732) -- (axis cs:66.3815,-16.3732);
      \node[font=\fontsize{5.6}{6.4}\selectfont, text=black, anchor=center, fill=white, inner sep=0.6pt] at (axis cs:66.3815,-16.3732) {qwen-2.5-72b};
      \draw[gray!60, line width=0.3pt] (axis cs:12.2934,-4.5243) -- (axis cs:12.2934,23.9591);
      \node[font=\fontsize{5.6}{6.4}\selectfont, text=black, anchor=center, fill=white, inner sep=0.6pt] at (axis cs:12.2934,23.9591) {qwen3.5-397b};
      \draw[gray!60, line width=0.3pt] (axis cs:-36.8546,43.9386) -- (axis cs:-36.8546,66.0924);
      \node[font=\fontsize{5.6}{6.4}\selectfont, text=black, anchor=center, fill=white, inner sep=0.6pt] at (axis cs:-36.8546,66.0924) {gpt-5.4-mini};
      \draw[gray!60, line width=0.3pt] (axis cs:44.1936,-2.7490) -- (axis cs:73.2573,-2.7490);
      \node[font=\fontsize{5.6}{6.4}\selectfont, text=black, anchor=center, fill=white, inner sep=0.6pt] at (axis cs:73.2573,-2.7490) {gemma-3-12b};
      \draw[gray!60, line width=0.3pt] (axis cs:46.6350,2.7476) -- (axis cs:57.6446,11.1396);
      \node[font=\fontsize{5.6}{6.4}\selectfont, text=black, anchor=center, fill=white, inner sep=0.6pt] at (axis cs:57.6446,11.1396) {gemma-3-27b};
      \draw[gray!60, line width=0.3pt] (axis cs:-11.1189,-14.1490) -- (axis cs:-26.6887,-14.1490);
      \node[font=\fontsize{5.6}{6.4}\selectfont, text=black, anchor=center, fill=white, inner sep=0.6pt] at (axis cs:-26.6887,-14.1490) {minimax-m3};
      \draw[gray!60, line width=0.3pt] (axis cs:-18.4294,12.8358) -- (axis cs:-18.4294,41.3193);
      \node[font=\fontsize{5.6}{6.4}\selectfont, text=black, anchor=center, fill=white, inner sep=0.6pt] at (axis cs:-18.4294,41.3193) {kimi-k2.6};
      \draw[gray!60, line width=0.3pt] (axis cs:-22.2548,7.1923) -- (axis cs:-22.2548,-21.2911);
      \node[font=\fontsize{5.6}{6.4}\selectfont, text=black, anchor=center, fill=white, inner sep=0.6pt] at (axis cs:-22.2548,-21.2911) {kimi-k2.5};
      \draw[gray!60, line width=0.3pt] (axis cs:-48.0956,25.2472) -- (axis cs:-48.0956,17.3352);
      \node[font=\fontsize{5.6}{6.4}\selectfont, text=black, anchor=center, fill=white, inner sep=0.6pt] at (axis cs:-48.0956,17.3352) {grok-4.3};
      \draw[gray!60, line width=0.3pt] (axis cs:-4.9664,9.7177) -- (axis cs:21.4565,29.8585);
      \node[font=\fontsize{5.6}{6.4}\selectfont, text=black, anchor=center, fill=white, inner sep=0.6pt] at (axis cs:21.4565,29.8585) {glm-4.7};
      \draw[gray!60, line width=0.3pt] (axis cs:-42.1474,39.6493) -- (axis cs:-57.7172,39.6493);
      \node[font=\fontsize{5.6}{6.4}\selectfont, text=black, anchor=center, fill=white, inner sep=0.6pt] at (axis cs:-57.7172,39.6493) {gpt-5.4};
      \draw[gray!60, line width=0.3pt] (axis cs:3.7250,2.3623) -- (axis cs:14.1049,2.3623);
      \node[font=\fontsize{5.6}{6.4}\selectfont, text=black, anchor=center, fill=white, inner sep=0.6pt] at (axis cs:14.1049,2.3623) {glm-5};
    \nextgroupplot[title={$k$-means++}]
      \addplot[only marks, mark=*, mark size=1.7pt, color=cluster0, mark options={fill=cluster0, draw=white, line width=0.4pt}] coordinates {(37.6480,-30.8920) (37.1715,-24.1102) (44.1936,-2.7490) (46.6350,2.7476) (52.5495,-14.3696) (55.7034,-8.8385) (12.2934,-4.5243) (-4.1004,-22.6952) (0.7373,-18.9073) (20.2308,-5.2400) (21.4558,-12.6927)};
      \addplot[only marks, mark=*, mark size=2.6pt, color=cluster0, mark options={fill=cluster0, draw=white, line width=0.4pt}] coordinates {(29.0138,-16.3732)};
      \addplot[only marks, mark=*, mark size=1.7pt, color=cluster1, mark options={fill=cluster1, draw=white, line width=0.4pt}] coordinates {(-11.1189,-14.1490) (-29.6915,8.3274) (-22.2548,7.1923) (-22.9018,21.4683) (-4.9664,9.7177) (3.7250,2.3623) (-31.4614,19.1747)};
      \addplot[only marks, mark=*, mark size=2.6pt, color=cluster1, mark options={fill=cluster1, draw=white, line width=0.4pt}] coordinates {(-18.4294,12.8358)};
      \addplot[only marks, mark=*, mark size=1.7pt, color=cluster2, mark options={fill=cluster2, draw=white, line width=0.4pt}] coordinates {(-48.0956,25.2472) (-41.3660,31.5803) (-29.2996,48.2286) (-36.8546,43.9386)};
      \addplot[only marks, mark=*, mark size=2.6pt, color=cluster2, mark options={fill=cluster2, draw=white, line width=0.4pt}] coordinates {(-42.1474,39.6493)};
      \draw[gray!60, line width=0.3pt] (axis cs:-29.2996,48.2286) -- (axis cs:-29.2996,56.1407);
      \node[font=\fontsize{5.6}{6.4}\selectfont, text=black, anchor=center, fill=white, inner sep=0.6pt] at (axis cs:-29.2996,56.1407) {nemotron-3-super-120b};
      \draw[gray!60, line width=0.3pt] (axis cs:-31.4614,19.1747) -- (axis cs:-31.4614,-2.9791);
      \node[font=\fontsize{5.6}{6.4}\selectfont, text=black, anchor=center, fill=white, inner sep=0.6pt] at (axis cs:-31.4614,-2.9791) {nemotron-3-ultra-550b};
      \draw[gray!60, line width=0.3pt] (axis cs:20.2308,-5.2400) -- (axis cs:20.2308,16.9138);
      \node[font=\fontsize{5.6}{6.4}\selectfont, text=black, anchor=center, fill=white, inner sep=0.6pt] at (axis cs:20.2308,16.9138) {gemini-3.1-flash-lite};
      \draw[gray!60, line width=0.3pt] (axis cs:52.5495,-14.3696) -- (axis cs:63.5590,-22.7616);
      \node[font=\fontsize{5.6}{6.4}\selectfont, text=black, anchor=center, fill=white, inner sep=0.6pt] at (axis cs:63.5590,-22.7616) {mistral-large-2512};
      \draw[gray!60, line width=0.3pt] (axis cs:55.7034,-8.8385) -- (axis cs:55.7034,-37.3219);
      \node[font=\fontsize{5.6}{6.4}\selectfont, text=black, anchor=center, fill=white, inner sep=0.6pt] at (axis cs:55.7034,-37.3219) {mistral-medium-3.1};
      \draw[gray!60, line width=0.3pt] (axis cs:-22.9018,21.4683) -- (axis cs:-11.8923,29.8603);
      \node[font=\fontsize{5.6}{6.4}\selectfont, text=black, anchor=center, fill=white, inner sep=0.6pt] at (axis cs:-11.8923,29.8603) {deepseek-chat-v3.1};
      \draw[gray!60, line width=0.3pt] (axis cs:0.7373,-18.9073) -- (axis cs:0.7373,-30.7754);
      \node[font=\fontsize{5.6}{6.4}\selectfont, text=black, anchor=center, fill=white, inner sep=0.6pt] at (axis cs:0.7373,-30.7754) {claude-sonnet-4.6};
      \draw[gray!60, line width=0.3pt] (axis cs:37.6480,-30.8920) -- (axis cs:37.6480,-47.5074);
      \node[font=\fontsize{5.6}{6.4}\selectfont, text=black, anchor=center, fill=white, inner sep=0.6pt] at (axis cs:37.6480,-47.5074) {llama-4-maverick};
      \draw[gray!60, line width=0.3pt] (axis cs:-4.1004,-22.6952) -- (axis cs:-4.1004,-39.3105);
      \node[font=\fontsize{5.6}{6.4}\selectfont, text=black, anchor=center, fill=white, inner sep=0.6pt] at (axis cs:-4.1004,-39.3105) {claude-haiku-4.5};
      \draw[gray!60, line width=0.3pt] (axis cs:21.4558,-12.6927) -- (axis cs:21.4558,9.4611);
      \node[font=\fontsize{5.6}{6.4}\selectfont, text=black, anchor=center, fill=white, inner sep=0.6pt] at (axis cs:21.4558,9.4611) {gemini-3.1-pro};
      \draw[gray!60, line width=0.3pt] (axis cs:37.1715,-24.1102) -- (axis cs:15.3737,-24.1102);
      \node[font=\fontsize{5.6}{6.4}\selectfont, text=black, anchor=center, fill=white, inner sep=0.6pt] at (axis cs:15.3737,-24.1102) {llama-3.3-70b};
      \draw[gray!60, line width=0.3pt] (axis cs:-41.3660,31.5803) -- (axis cs:-61.9172,47.2454);
      \node[font=\fontsize{5.6}{6.4}\selectfont, text=black, anchor=center, fill=white, inner sep=0.6pt] at (axis cs:-61.9172,47.2454) {deepseek-v3.2};
      \draw[gray!60, line width=0.3pt] (axis cs:-29.6915,8.3274) -- (axis cs:-51.4893,8.3274);
      \node[font=\fontsize{5.6}{6.4}\selectfont, text=black, anchor=center, fill=white, inner sep=0.6pt] at (axis cs:-51.4893,8.3274) {minimax-m2.5};
      \draw[gray!60, line width=0.3pt] (axis cs:29.0138,-16.3732) -- (axis cs:66.3815,-16.3732);
      \node[font=\fontsize{5.6}{6.4}\selectfont, text=black, anchor=center, fill=white, inner sep=0.6pt] at (axis cs:66.3815,-16.3732) {qwen-2.5-72b};
      \draw[gray!60, line width=0.3pt] (axis cs:12.2934,-4.5243) -- (axis cs:12.2934,23.9591);
      \node[font=\fontsize{5.6}{6.4}\selectfont, text=black, anchor=center, fill=white, inner sep=0.6pt] at (axis cs:12.2934,23.9591) {qwen3.5-397b};
      \draw[gray!60, line width=0.3pt] (axis cs:-36.8546,43.9386) -- (axis cs:-36.8546,66.0924);
      \node[font=\fontsize{5.6}{6.4}\selectfont, text=black, anchor=center, fill=white, inner sep=0.6pt] at (axis cs:-36.8546,66.0924) {gpt-5.4-mini};
      \draw[gray!60, line width=0.3pt] (axis cs:44.1936,-2.7490) -- (axis cs:73.2573,-2.7490);
      \node[font=\fontsize{5.6}{6.4}\selectfont, text=black, anchor=center, fill=white, inner sep=0.6pt] at (axis cs:73.2573,-2.7490) {gemma-3-12b};
      \draw[gray!60, line width=0.3pt] (axis cs:46.6350,2.7476) -- (axis cs:57.6446,11.1396);
      \node[font=\fontsize{5.6}{6.4}\selectfont, text=black, anchor=center, fill=white, inner sep=0.6pt] at (axis cs:57.6446,11.1396) {gemma-3-27b};
      \draw[gray!60, line width=0.3pt] (axis cs:-11.1189,-14.1490) -- (axis cs:-26.6887,-14.1490);
      \node[font=\fontsize{5.6}{6.4}\selectfont, text=black, anchor=center, fill=white, inner sep=0.6pt] at (axis cs:-26.6887,-14.1490) {minimax-m3};
      \draw[gray!60, line width=0.3pt] (axis cs:-18.4294,12.8358) -- (axis cs:-18.4294,41.3193);
      \node[font=\fontsize{5.6}{6.4}\selectfont, text=black, anchor=center, fill=white, inner sep=0.6pt] at (axis cs:-18.4294,41.3193) {kimi-k2.6};
      \draw[gray!60, line width=0.3pt] (axis cs:-22.2548,7.1923) -- (axis cs:-22.2548,-21.2911);
      \node[font=\fontsize{5.6}{6.4}\selectfont, text=black, anchor=center, fill=white, inner sep=0.6pt] at (axis cs:-22.2548,-21.2911) {kimi-k2.5};
      \draw[gray!60, line width=0.3pt] (axis cs:-48.0956,25.2472) -- (axis cs:-48.0956,17.3352);
      \node[font=\fontsize{5.6}{6.4}\selectfont, text=black, anchor=center, fill=white, inner sep=0.6pt] at (axis cs:-48.0956,17.3352) {grok-4.3};
      \draw[gray!60, line width=0.3pt] (axis cs:-4.9664,9.7177) -- (axis cs:21.4565,29.8585);
      \node[font=\fontsize{5.6}{6.4}\selectfont, text=black, anchor=center, fill=white, inner sep=0.6pt] at (axis cs:21.4565,29.8585) {glm-4.7};
      \draw[gray!60, line width=0.3pt] (axis cs:-42.1474,39.6493) -- (axis cs:-57.7172,39.6493);
      \node[font=\fontsize{5.6}{6.4}\selectfont, text=black, anchor=center, fill=white, inner sep=0.6pt] at (axis cs:-57.7172,39.6493) {gpt-5.4};
      \draw[gray!60, line width=0.3pt] (axis cs:3.7250,2.3623) -- (axis cs:14.1049,2.3623);
      \node[font=\fontsize{5.6}{6.4}\selectfont, text=black, anchor=center, fill=white, inner sep=0.6pt] at (axis cs:14.1049,2.3623) {glm-5};
    \end{groupplot}
  \end{tikzpicture}
  \caption{Both panels place the 25 models at identical t-SNE coordinates; Larger marks are medoids. Note that the hierarchical clustering result is unevenly sized: grok-4.3 is the only member of a cluster. The k-means++ results are stable with a stability score of 0.767.}
  \label{fig:groups-tsne}
\end{figure*}

We further assess the robustness of the identified clustering structure through a resampling-based stability analysis, following Hennig et al. \cite{hennig2007}. Specifically, we generate 500 bootstrap replicates by resampling the 350 development questions with replacement. For each replicate, we reconstruct the model representations and repeat the complete clustering procedure using the same configuration. Each cluster in the original solution is matched to the replicate cluster with which it has the highest Jaccard similarity. We then average the Jaccard similarities across clusters and bootstrap replicates to obtain the stability score. This analysis evaluates whether the inferred grouping of models is robust to variation in the particular questions used to construct their representations, rather than being specific to a single development sample. At $K=3$, the resulting stability score is 0.767, indicating reasonably stable cluster membership under bootstrap perturbations \cite{hennig2007}. Together with its silhouette score, this result provides evidence that the $K=3$ solution captures a reproducible clustering structure rather than one that is highly sensitive to the sampled development questions.

\paragraph{The choice of $K$ and $k$}
\label{sec:choice_of_k}

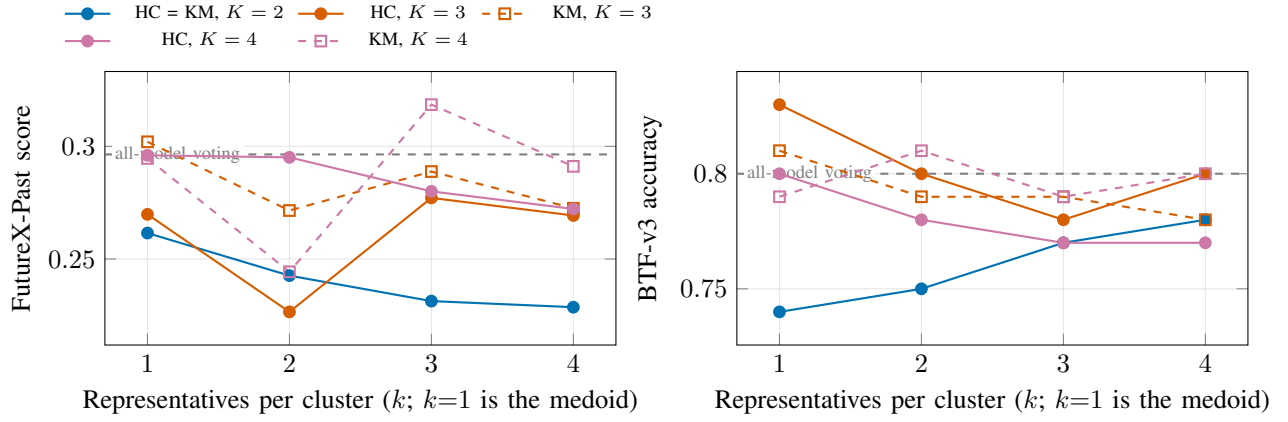
\begin{figure*}[t]
\centering
\definecolor{swBlue}{RGB}{0,114,178}
\definecolor{swPurple}{RGB}{204,121,167}
\definecolor{swVermillion}{RGB}{213,94,0}
\begin{tikzpicture}
\begin{groupplot}[
  group style={group size=2 by 1, horizontal sep=1.6cm},
  width=0.46\textwidth, height=5.2cm,
  xlabel={Representatives per cluster ($k$; $k{=}1$ is the medoid)},
  xtick={1,2,3,4}, grid=both, grid style={gray!20},
  legend style={font=\scriptsize, at={(0.5,1.04)}, anchor=south, legend columns=3,
                draw=none, fill=none, column sep=0.7em, /tikz/every even column/.append style={column sep=0.3em}},
]
\nextgroupplot[ylabel={FutureX-Past score}, xmin=0.7, xmax=4.3,
  ymin=0.2118, ymax=0.3332]
\addplot[gray, dashed, thick, forget plot] coordinates {(0.7,0.2964) (4.3,0.2964)};
\node[font=\scriptsize, gray, anchor=west, fill=white, inner sep=1pt] at (axis cs:0.75,0.2964) {all-model voting};
\addplot[color=swBlue, solid, thick, mark=*, mark size=2pt, mark options={solid, fill=swBlue}] coordinates {(1,0.2615) (2,0.2426) (3,0.2313) (4,0.2286)};
\addlegendentry{HC = KM, $K=2$}
\addplot[color=swVermillion, solid, thick, mark=*, mark size=2pt, mark options={solid, fill=swVermillion}] coordinates {(1,0.2698) (2,0.2265) (3,0.2771) (4,0.2693)};
\addlegendentry{HC, $K=3$}
\addplot[color=swVermillion, dashed, thick, mark=square, mark size=2pt, mark options={solid, fill=swVermillion}] coordinates {(1,0.302) (2,0.2715) (3,0.2888) (4,0.2725)};
\addlegendentry{KM, $K=3$}
\addplot[color=swPurple, solid, thick, mark=*, mark size=2pt, mark options={solid, fill=swPurple}] coordinates {(1,0.296) (2,0.2951) (3,0.28) (4,0.2722)};
\addlegendentry{HC, $K=4$}
\addplot[color=swPurple, dashed, thick, mark=square, mark size=2pt, mark options={solid, fill=swPurple}] coordinates {(1,0.2946) (2,0.2443) (3,0.3185) (4,0.2911)};
\addlegendentry{KM, $K=4$}
\nextgroupplot[ylabel={BTF-v3 accuracy}, xmin=0.7, xmax=4.3,
  ymin=0.7256, ymax=0.8444]
\addplot[gray, dashed, thick, forget plot] coordinates {(0.7,0.8) (4.3,0.8)};
\node[font=\scriptsize, gray, anchor=west, fill=white, inner sep=1pt] at (axis cs:0.75,0.8) {all-model voting};
\addplot[color=swBlue, solid, thick, mark=*, mark size=2pt, mark options={solid, fill=swBlue}, forget plot] coordinates {(1,0.74) (2,0.75) (3,0.77) (4,0.78)};
\addplot[color=swVermillion, solid, thick, mark=*, mark size=2pt, mark options={solid, fill=swVermillion}, forget plot] coordinates {(1,0.83) (2,0.8) (3,0.78) (4,0.8)};
\addplot[color=swVermillion, dashed, thick, mark=square, mark size=2pt, mark options={solid, fill=swVermillion}, forget plot] coordinates {(1,0.81) (2,0.79) (3,0.79) (4,0.78)};
\addplot[color=swPurple, solid, thick, mark=*, mark size=2pt, mark options={solid, fill=swPurple}, forget plot] coordinates {(1,0.8) (2,0.78) (3,0.77) (4,0.77)};
\addplot[color=swPurple, dashed, thick, mark=square, mark size=2pt, mark options={solid, fill=swPurple}, forget plot] coordinates {(1,0.79) (2,0.81) (3,0.79) (4,0.8)};
\end{groupplot}
\end{tikzpicture}
\caption{Score against the number of representatives drawn from each cluster. Colour is the cluster count $K$, line style the clustering algorithm (solid HC~=~hierarchical, dashed KM~=~$K$-means++); a label of the form \emph{HC~=~KM} marks two clusterings that produce identical committees at every $k$ and are therefore drawn once. $k{=}1$ is the cluster medoid; $k\ge2$ takes the $k$ most mutually dissimilar members, and all representatives are pooled into a single vote. The grey dashed line is all-model voting over the full 25-model crowd. Scores are over 100 held-out items from a seeded five-fold split. Clusterings with no dissimilar set defined beyond $k{=}2$ stop there.}
\label{fig:k_sweep}
\end{figure*}

Two integers define a crowd: the number of clusters $K$, and the number of representatives $k$ drawn
from each. Figure~\ref{fig:k_sweep} sweeps both. It also explains a choice that the clustering metrics
alone would have decided differently.

\textbf{Why not $K=2$, which the metric prefers.}
By silhouette score, $K=2$ is the best partition of the 25 models: 0.334 against 0.237 at $K=3$
(Figure~\ref{fig:best-k}). On the internal clustering metric it wins outright. It is nonetheless a poor
choice here, for the reason noted above: two clusters yield two medoid representatives,
and a similarity medoid over two candidates is tied by construction. That risk is measurable by the tie
analysis of Appendix~\ref{sec:tiebreak}, and the cost is large. At $K=2$, \textbf{72 of the
100 FutureX-Past questions} are decided by a tie-break rather than by the aggregation rule, and the
score consequently ranges over $0.209$ depending on which convention resolves them --- a spread wider
than any difference this paper reports. At $K=3$ the same measurement gives 28 tied questions under
hierarchical clustering and 32 under $K$-means++, with ranges of $0.012$ and $0.052$. Adding one cluster
converts most decisions from arbitration back into aggregation. That matters because
Appendix~\ref{sec:tiebreak} finds no deterministic tie-break that beats random resolution: the fewer
decisions a tie-break has to make, the less of the reported score rests on a rule that carries no
information about which answer is correct.

Prediction performance agrees. In Figure~\ref{fig:k_sweep} the $K=2$ curve is the lowest on FutureX-Past
at every $k$ and declines monotonically as $k$ grows (0.262, 0.243, 0.231, 0.229), whereas $K=3$ under
$K$-means++ starts at 0.302 and stays above all-model voting's 0.296 at $k=1$. A partition that scores
best on cluster separation therefore produces the weakest crowd, which is the central caution of this
subsection: \emph{internal clustering quality is not a proxy for downstream crowd quality}, and selecting
$K$ on silhouette alone would have chosen the worst available option.

\textbf{Why $k=1$.}
Across both clusterings and every $K$, drawing more representatives per cluster does not
systematically help. The trend is uneven rather than monotone --- $K=4$ under $K$-means++ peaks at $k=3$
(0.319), while $K=3$ under $K$-means++ peaks at $k=1$ (0.302) --- and no setting with $k>1$ beats the
best $k=1$ setting on FutureX-Past. Since cost grows linearly in $k$ while the score does not, $k=1$ (the
medoid) is both the cheapest and the strongest choice. We therefore fix $k=1$ throughout, which makes the
behaviour-aware crowd both the cheapest and the strongest configuration we measured.

With the best $k$ determined, we run $K$-means++ and hierarchical clustering, respectively, to generate alternative candidate diverse groups. We then visualize the 25 model representations in a two-dimensional projection with t-SNE \cite{maaten08} (Figure \ref{fig:groups-tsne}) . 



\subsection{RQ2: Diversity-Aware Future Prediction}
\label{sec:rq2}

We next investigate whether the behavioral diversity discovered on the development benchmarks transfers to the distinct task of future prediction.

Importantly, the behavioral clusters and representatives are determined without using any of the future-prediction questions. Thus, this experiment evaluates whether diversity inferred from general reasoning behavior provides useful information for constructing a crowd in a
different prediction domain.

We compare individual models, conventional all-model voting, performance-based expert selection, random representative voting, and the three behavior-aware crowd strategies.

\begin{table*}[t]
\caption{Future-prediction performance of individual and crowd-based methods, over 100 held-out items from a seeded five-fold split, pooled so that every question contributes exactly once. Per-benchmark rank in parentheses; \textbf{Avg. Rank} is the mean of the two ranks rather than of the scores, because the two metrics are on different scales. HC~=~hierarchical clustering, KM~=~$K$-means++, both at $K=3$. The best value in each benchmark column is shown in bold.}
\label{tab:prediction_results}
\centering
\small
\begin{tabular}{lcccc}
\toprule
Method & FutureX-Past & BTF-v3 & Avg. Rank & \# Models \\
\midrule
Performance-based expert           & 0.265 (10)           & 0.790 (7)            & 8.5 & 1       \\
All-model voting                   & 0.296 (2)            & 0.800 (4)            & 3.0 & 25      \\
\midrule
Random representatives (KM)        & 0.274 (6)            & 0.771 (10)           & 8.0 & 3       \\
Random representatives (HC)        & 0.285 (3)            & 0.798 (6)            & 4.5 & 3       \\
\addlinespace
Medoid voting (KM)                 & \textbf{0.302 (1)}   & 0.810 (2)            & 1.5 & 3       \\
Medoid voting (HC)                 & 0.270 (8)            & \textbf{0.830 (1)}   & 4.5 & 3       \\
\addlinespace
$k=4$ dissimilar voting (KM)       & 0.272 (7)            & 0.780 (9)            & 8.0 & 12      \\
$k=4$ dissimilar voting (HC)       & 0.269 (9)            & 0.800 (4)            & 6.5 & 9       \\
\addlinespace
Cluster subcommittee voting (KM)   & 0.284 (4)            & 0.790 (7)            & 5.5 & 25      \\
Cluster subcommittee voting (HC)   & 0.282 (5)            & 0.810 (2)            & 3.5 & 25      \\
\bottomrule
\end{tabular}
\end{table*}

Table~\ref{tab:prediction_results} reports the pooled held-out results. Medoid voting under the
$K$-means++ clustering attains the highest combined rank, exceeding all-model voting on both
benchmarks with $0.302$ on FutureX-Past and $0.810$ on BTF-v3 while querying $K=3$ models rather than
all $25$. The medoids in this best setting include qwen-2.5-72b-instruct of the 12-member cluster, kimi-k2.6 of the 8-member cluster, and gpt-5.4 of the 5-member cluster.

All-model voting follows at $0.296$ and $0.800$, and the performance-based
expert---the only method that consumes future-prediction labels---reaches $0.265$ and $0.790$. The
hierarchical medoid crowd takes the single best cell in the table, $0.830$ on BTF-v3, but falls to
eighth on FutureX-Past at $0.270$:

The comparison against size-matched random representatives isolates the contribution of
behavior-aware selection. Drawing one model per cluster at random $1000$ times, the $K$-means++
medoid line-up scores at the $88$th percentile of that distribution on BTF-v3 ($829$ random line-ups
worse, $92$ identical, $79$ better) and the $83$rd on FutureX-Past ($833$/$2$/$165$). Crowds of the
same size drawn at random therefore match the medoid crowd in roughly one draw in eight.

Two qualifications follow from the table. First, the effect belongs to the $K$-means++ partition
rather than to medoids in general: under hierarchical clustering the same construction reaches
$0.270$ on FutureX-Past, eighth of ten, although it takes BTF-v3 with $0.830$. Second, the two
behavior-aware variants do not both transfer. Against its own size-matched control, $k=4$ dissimilar
voting sits at the $5$th percentile on FutureX-Past under $K$-means++ and the $8$th under
hierarchical clustering---worse than $947$ and $917$ of $1000$ random line-ups of the same
size---while reaching only the $32$nd and $61$st percentiles on BTF-v3. Maximizing
within-cluster distance therefore selects models that disagree without being individually better,
and on a benchmark scored by overlap that discards more correct answers than it filters incorrect
ones. Cluster subcommittee voting scores below plain all-model voting on FutureX-Past, giving no
evidence that equalizing cluster influence helps once all models already participate.

The comparison with all-model voting tests whether explicitly controlling behavioral redundancy improves conventional crowd aggregation. The comparison with random subsets determines whether any gain from representative voting can be attributed specifically to
behavior-aware selection rather than simply to using fewer models.

To further separate the benefit of behavioral clustering from the specific representative-selection criterion, we optionally compare $k$-dissimilar representatives against randomly selected representatives within each cluster.
However, the result shows that maximum diversity does not necessarily help the overall performance. Rather, it delivers much worse performance than medoid voting.

\paragraph{Two questions this raises, answered in the appendices}
The results above invite two follow-up questions, and we report both in full after the references rather
than interrupt the cost analysis with them.

The first is whether behavioral diversity is what carries the gain, or whether the constructed crowds
merely happen to be good ones. Appendix~\ref{sec:when_diversity_helps} answers this at the population
level, over 498 candidate committees rather than the handful compared here. Holding committee size fixed
--- which is essential, because size alone correlates $+0.505$ with the FutureX-Past score and washes out
everything else --- a committee's internal behavioral disagreement still tracks its score at $+0.211$,
positive in all seven size strata. The effect is absent on BTF-v3, which the aggregation rules predict:
a median over probabilities averages disagreement away before it can act. Diversity is therefore
associated with stronger crowds, though not strongly enough to select on directly.

The second is how much of the FutureX-Past ordering is decided by the tie-break rather than by the
methods. Appendix~\ref{sec:tiebreak} measures this. Ties are frequent and structural, rising from
$22.3\%$ of decisions at level~1 to $80.6\%$ at level~4, and no deterministic rule we tested beats random
resolution. The alphabetical rule behind the numbers reported here is in fact the unlucky end of that
range, so the differences above were obtained under a convention that works against them rather than for
them.  

\subsection{RQ3: Prediction Accuracy versus Crowd Cost}
\label{sec:rq3}

Finally, we investigate whether behavioral diversity can reduce the costs required for effective crowd prediction.

All-model voting requires 25 model calls for each prediction. In contrast, medoid voting requires approximately $K$ calls, while the $k$-representative strategy requires approximately $kK$
calls, subject to cluster size. Cluster-subcommittee voting retains all 25 models and therefore serves primarily as a diversity-aware
aggregation method rather than a cost-reduction strategy.

For each representative method, we calculate the model-call reduction relative to all-model voting.
The primary analysis plots future-prediction accuracy against the number of models queried per question.

\begin{table*}[t]
\caption{Cost of each method per 100 questions, against the score it achieves. Calls are the queries a method commits to, $m$ models $\times$ 100 questions, whether or not every one returns a usable answer. Dollars are the amounts the provider actually charged, not a price-times-tokens estimate. \emph{Rel.} is cost relative to all-model voting on the same benchmark. $K$-means++ clustering, $K=3$.}
\label{tab:cost}
\centering
\begin{tabular}{lrrrrr}
\toprule
& & \multicolumn{2}{c}{FutureX-Past} & \multicolumn{2}{c}{BTF-v3} \\
\cmidrule(lr){3-4}\cmidrule(lr){5-6}
Method & Calls & USD (rel.) & Score & USD (rel.) & Score \\
\midrule
Performance-based expert       &   100 &  0.11 (0.01) & 0.265 &  1.31 (0.06) & 0.790 \\
All-model voting               &  2500 & 10.96 (1.00) & 0.296 & 21.26 (1.00) & 0.800 \\
Random representatives         &   300 &  1.25 (0.11) & 0.274 &  2.41 (0.11) & 0.771 \\
Medoid voting                  &   300 &  2.10 (0.19) & 0.302 &  4.51 (0.21) & 0.810 \\
$k=4$ dissimilar voting        &  1200 &  7.12 (0.65) & 0.272 & 13.91 (0.65) & 0.780 \\
Cluster subcommittee voting    &  2500 & 10.96 (1.00) & 0.284 & 21.26 (1.00) & 0.790 \\
\bottomrule
\end{tabular}
\end{table*}

We compare medoid and $k$-dissimilar selection with random subsets of the same size. This controls for the possibility that reducing the crowd size alone, rather than preserving behavioral diversity, accounts
for the observed performance.

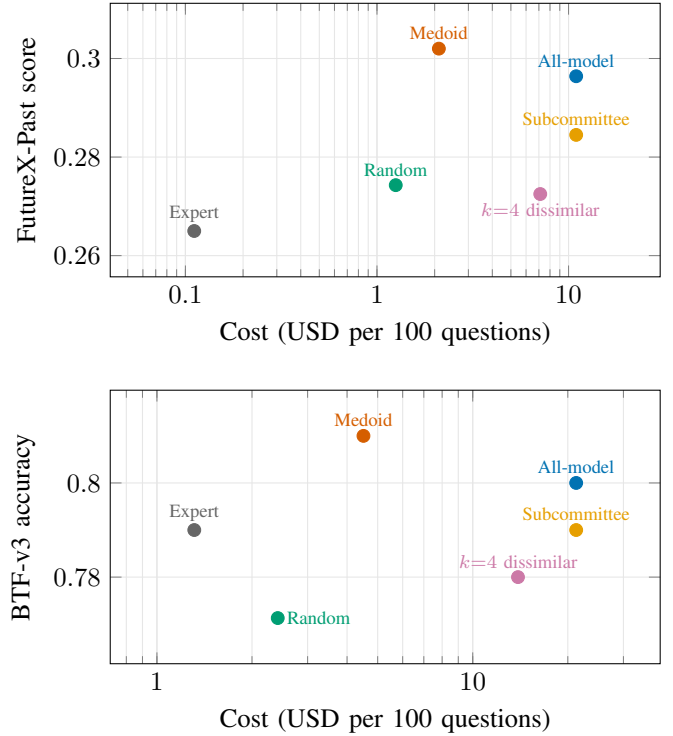
\begin{figure}[t]
\centering
\definecolor{figBlue}{RGB}{0,114,178}
\definecolor{figGreen}{RGB}{0,158,115}
\definecolor{figGrey}{RGB}{102,102,102}
\definecolor{figOrange}{RGB}{230,159,0}
\definecolor{figPurple}{RGB}{204,121,167}
\definecolor{figVermillion}{RGB}{213,94,0}
\begin{tikzpicture}
\begin{groupplot}[
  group style={group size=1 by 2, vertical sep=1.5cm},
  width=\columnwidth, height=5.2cm,
  xlabel={Cost (USD per 100 questions)}, grid=both, grid style={gray!20},
  log ticks with fixed point,
]
\nextgroupplot[ylabel={FutureX-Past score}, xmode=log,
  xmin=0.04052, xmax=30.08,
  ymin=0.2557, ymax=0.3112]
\addplot[only marks, mark=*, mark size=2.4pt, color=figBlue] coordinates {(10.96,0.2964)};
\addplot[only marks, mark=*, mark size=2.4pt, color=figGrey] coordinates {(0.1112,0.265)};
\addplot[only marks, mark=*, mark size=2.4pt, color=figGreen] coordinates {(1.253,0.2743)};
\addplot[only marks, mark=*, mark size=2.4pt, color=figVermillion] coordinates {(2.104,0.302)};
\addplot[only marks, mark=*, mark size=2.4pt, color=figPurple] coordinates {(7.116,0.2725)};
\addplot[only marks, mark=*, mark size=2.4pt, color=figOrange] coordinates {(10.96,0.2845)};
\node[font=\scriptsize, text=figBlue, above] at (axis cs:10.96,0.2964) {All-model};
\node[font=\scriptsize, text=figGrey, above] at (axis cs:0.1112,0.265) {Expert};
\node[font=\scriptsize, text=figGreen, above] at (axis cs:1.253,0.2743) {Random};
\node[font=\scriptsize, text=figVermillion, above] at (axis cs:2.104,0.302) {Medoid};
\node[font=\scriptsize, text=figPurple, below] at (axis cs:7.116,0.2725) {$k{=}4$ dissimilar};
\node[font=\scriptsize, text=figOrange, above] at (axis cs:10.96,0.2845) {Subcommittee};
\nextgroupplot[ylabel={BTF-v3 accuracy}, xmode=log,
  xmin=0.7115, xmax=39.23,
  ymin=0.7616, ymax=0.8197]
\addplot[only marks, mark=*, mark size=2.4pt, color=figBlue] coordinates {(21.26,0.8)};
\addplot[only marks, mark=*, mark size=2.4pt, color=figGrey] coordinates {(1.313,0.79)};
\addplot[only marks, mark=*, mark size=2.4pt, color=figGreen] coordinates {(2.413,0.7713)};
\addplot[only marks, mark=*, mark size=2.4pt, color=figVermillion] coordinates {(4.508,0.81)};
\addplot[only marks, mark=*, mark size=2.4pt, color=figPurple] coordinates {(13.91,0.78)};
\addplot[only marks, mark=*, mark size=2.4pt, color=figOrange] coordinates {(21.26,0.79)};
\node[font=\scriptsize, text=figBlue, above] at (axis cs:21.26,0.8) {All-model};
\node[font=\scriptsize, text=figGrey, above] at (axis cs:1.313,0.79) {Expert};
\node[font=\scriptsize, text=figGreen, right] at (axis cs:2.413,0.7713) {Random};
\node[font=\scriptsize, text=figVermillion, above] at (axis cs:4.508,0.81) {Medoid};
\node[font=\scriptsize, text=figPurple, above] at (axis cs:13.91,0.78) {$k{=}4$ dissimilar};
\node[font=\scriptsize, text=figOrange, above] at (axis cs:21.26,0.79) {Subcommittee};
\end{groupplot}
\end{tikzpicture}
\caption{Future-prediction score against measured cost in USD, for the $K$-means++ clustering at $K=3$. Medoid voting sits above and to the left of all-model voting on both benchmarks: a higher score for about 20\% of the outlay. Costs are the amounts the provider actually charged, summed over the 100 questions of each benchmark.}
\label{fig:accuracy_cost_usd}
\end{figure}

The central question is therefore not simply whether a smaller crowd can outperform all 25 models, but whether behavior-aware selection provides a more favorable \emph{accuracy--cost tradeoff} than conventional or randomly constructed crowds.


Table~\ref{tab:cost} reports what each method costs per 100 questions, measured as the amounts the
provider actually charged rather than estimated from list prices. Medoid voting scores above
all-model voting on both benchmarks while querying three models instead of 25, at $0.21\times$ the
cost on BTF-v3 (\$4.51 vs.\ \$21.26) and $0.19\times$ on FutureX-Past (\$2.10 vs.\
\$10.96). Cluster subcommittee voting costs exactly as much as all-model voting, since it queries
every model, and scores below it on both benchmarks: hierarchical vote weighting buys nothing once
the full crowd is already being paid for. 

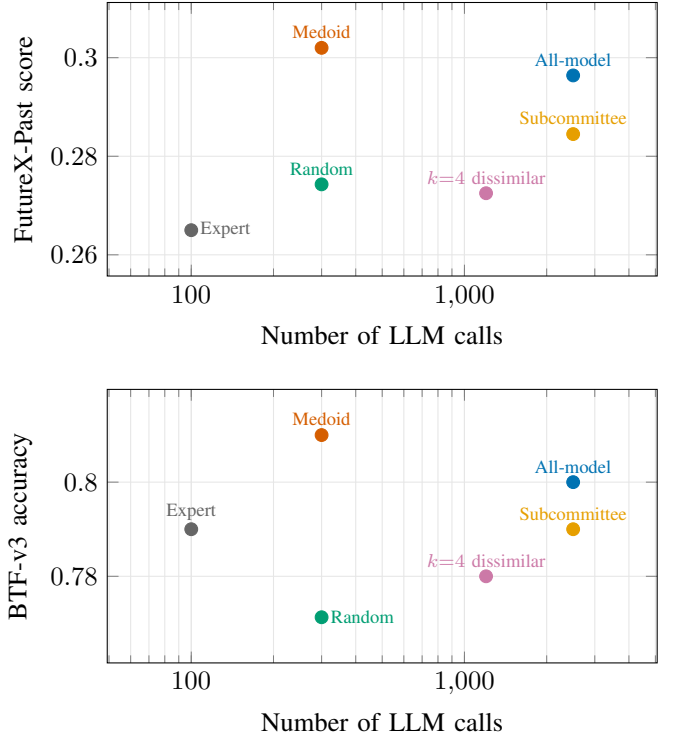
\begin{figure}[t]
\centering
\definecolor{figBlue}{RGB}{0,114,178}
\definecolor{figGreen}{RGB}{0,158,115}
\definecolor{figGrey}{RGB}{102,102,102}
\definecolor{figOrange}{RGB}{230,159,0}
\definecolor{figPurple}{RGB}{204,121,167}
\definecolor{figVermillion}{RGB}{213,94,0}
\begin{tikzpicture}
\begin{groupplot}[
  group style={group size=1 by 2, vertical sep=1.5cm},
  width=\columnwidth, height=5.2cm,
  xlabel={Number of LLM calls}, grid=both, grid style={gray!20},
  log ticks with fixed point,
]
\nextgroupplot[ylabel={FutureX-Past score}, xmode=log,
  xmin=49.26, xmax=5076,
  ymin=0.2557, ymax=0.3112]
\addplot[only marks, mark=*, mark size=2.4pt, color=figBlue] coordinates {(2500,0.2964)};
\addplot[only marks, mark=*, mark size=2.4pt, color=figGrey] coordinates {(100,0.265)};
\addplot[only marks, mark=*, mark size=2.4pt, color=figGreen] coordinates {(300,0.2743)};
\addplot[only marks, mark=*, mark size=2.4pt, color=figVermillion] coordinates {(300,0.302)};
\addplot[only marks, mark=*, mark size=2.4pt, color=figPurple] coordinates {(1200,0.2725)};
\addplot[only marks, mark=*, mark size=2.4pt, color=figOrange] coordinates {(2500,0.2845)};
\node[font=\scriptsize, text=figBlue, above] at (axis cs:2500,0.2964) {All-model};
\node[font=\scriptsize, text=figGrey, right] at (axis cs:100,0.265) {Expert};
\node[font=\scriptsize, text=figGreen, above] at (axis cs:300,0.2743) {Random};
\node[font=\scriptsize, text=figVermillion, above] at (axis cs:300,0.302) {Medoid};
\node[font=\scriptsize, text=figPurple, above] at (axis cs:1200,0.2725) {$k{=}4$ dissimilar};
\node[font=\scriptsize, text=figOrange, above] at (axis cs:2500,0.2845) {Subcommittee};
\nextgroupplot[ylabel={BTF-v3 accuracy}, xmode=log,
  xmin=49.26, xmax=5076,
  ymin=0.7616, ymax=0.8197]
\addplot[only marks, mark=*, mark size=2.4pt, color=figBlue] coordinates {(2500,0.8)};
\addplot[only marks, mark=*, mark size=2.4pt, color=figGrey] coordinates {(100,0.79)};
\addplot[only marks, mark=*, mark size=2.4pt, color=figGreen] coordinates {(300,0.7713)};
\addplot[only marks, mark=*, mark size=2.4pt, color=figVermillion] coordinates {(300,0.81)};
\addplot[only marks, mark=*, mark size=2.4pt, color=figPurple] coordinates {(1200,0.78)};
\addplot[only marks, mark=*, mark size=2.4pt, color=figOrange] coordinates {(2500,0.79)};
\node[font=\scriptsize, text=figBlue, above] at (axis cs:2500,0.8) {All-model};
\node[font=\scriptsize, text=figGrey, above] at (axis cs:100,0.79) {Expert};
\node[font=\scriptsize, text=figGreen, right] at (axis cs:300,0.7713) {Random};
\node[font=\scriptsize, text=figVermillion, above] at (axis cs:300,0.81) {Medoid};
\node[font=\scriptsize, text=figPurple, above] at (axis cs:1200,0.78) {$k{=}4$ dissimilar};
\node[font=\scriptsize, text=figOrange, above] at (axis cs:2500,0.79) {Subcommittee};
\end{groupplot}
\end{tikzpicture}
\caption{Future-prediction score against the number of LLM calls, for the $K$-means++ clustering at $K=3$. Medoid voting sits above and to the left of all-model voting on both benchmarks: a higher score for about 12\% of the calls. A method's call count is the queries it commits to, $m$ models $\times$ 100 questions, whether or not every one returns a usable answer.}
\label{fig:accuracy_cost_calls}
\end{figure}

Cost is dominated by model choice rather than crowd size. Per-call prices span roughly two orders of
magnitude across our population, and the single most expensive model costs more on its own
(\$6.46 per 100 questions on BTF-v3) than the entire three-model medoid committee.
Figures~\ref{fig:accuracy_cost_usd} and~\ref{fig:accuracy_cost_calls} plot score against measured
dollar cost and against the number of calls respectively; medoid voting sits above and to the left of
all-model voting on both benchmarks in both views.
Any crowd-construction method therefore trades on two axes at once, and reporting only the number of
calls understates the difference between crowds of equal size.

Two comparisons temper the efficiency claim. First, an oracle that could name the single best model
in advance would reach 0.820 on BTF-v3 for \$0.35, about one thirteenth of the
medoid committee's cost; two models tie at that score and \$0.35 is the cheaper of them. That model cannot be identified without labels: the realizable
single-model strategy is the performance-based expert, which reaches 0.790 for \$1.31. Medoid voting therefore falls $0.010$ short of that oracle while costing more, and
exceeds the realizable expert by $0.020$ accuracy at an additional \$3.20 per 100 questions,
without consulting any future-prediction label.

Second, restricting the crowd to open-weights models does not reduce cost in our population.
Voting over all 17 open-weights models costs \$9.66 and reaches 0.790 on BTF-v3,
against \$21.26 and 0.800 for the full crowd---cheaper, but at lower accuracy, and still far above
the cost of the cheapest single closed model. On this evidence, licensing is not a useful proxy for
inference cost.

\section{Conclusion}
This work investigates whether behavioral diversity can improve the
wisdom of LLM crowds for future prediction. We characterize LLMs using
their reasoning behavior on independent development tasks and use the
resulting behavioral groups to construct prediction crowds. Our results
show that crowd composition can matter more than crowd size. In
particular, a three-model medoid crowd based on $K$-means++ behavioral
clustering achieves the best performance on both future-prediction
benchmarks, outperforming conventional voting over all 25 models while
using only a small fraction of the model calls and the
inference cost. Size-matched random-crowd comparisons further support
the benefit of behavior-aware selection. 

Our results also suggest that useful crowd diversity is not simply a
matter of maximizing behavioral differences: maximally dissimilar
representatives and cluster-level voting do not consistently improve
prediction. A limitation of the current approach is that behavioral
signatures are constructed by averaging embeddings of observable
reasoning traces, which may obscure task-dependent behavioral patterns.
Moreover, the framework constructs a single static crowd, although model
complementarity may vary across prediction questions. Future work can
therefore explore richer behavioral representations and task-aware or
dynamically constructed crowds that jointly consider diversity,
representativeness, and predictive competence. 

Finally, our evaluation is limited to 25 LLMs and two future-prediction benchmarks with 100 sampled questions each. Future work will evaluate larger and evolving model populations across broader prediction domains to further validate the generality of these findings.
\section*{Acknowledgment} 
This material is based upon work partially supported by the U.S. National Science Foundation under Grant No. 2517121 and an UMBC Interdisciplinary Research fund.

\bibliographystyle{IEEEtran}
\bibliography{references,paper,llm} 

\appendices

\section{When Does Behavioral Diversity Help?}
\label{sec:when_diversity_helps}

The results above compare a small number of constructed crowds. They do not say whether behavioral
diversity itself is what carries the effect, or whether the constructed crowds simply happen to be
good. To separate these, we ask a population-level question: across many possible committees, does a
committee's \emph{internal behavioral disagreement} track how well it predicts?

We form 498 candidate committees over the 25 models---the structured ones used above plus seeded random
draws---spanning sizes 2 to 25. For each we measure two label-free quantities from the behavioral
representations alone: \emph{disagreement}, the mean pairwise behavioral distance among its members, and
\emph{consensus}, its complement. Neither uses any prediction outcome. We then correlate each against the
committee's benchmark score using a tie-aware Spearman coefficient computed as Pearson over competition
ranks, since scores over 100 items take many repeated values.

\subsection{Committee size must be held fixed}
Larger committees score better for reasons that have nothing to do with diversity, and larger committees
are also more internally diverse simply by having more members. Pooled across all sizes, size correlates
$+0.505$ with the FutureX-Past score. Any signal that grows with size will inherit that correlation. We
therefore stratify into seven size bands (24--83 committees each) and correlate within them. The
correction is not cosmetic: size itself falls from $+0.505$ to $+0.081$ once its own band is held fixed,
and \emph{coverage}---the union of behaviors a committee spans---falls from $+0.298$ to $+0.091$, which
identifies it as size in disguise. A pooled analysis would have credited several signals that do not
survive the control.

\subsection{Disagreement survives the control}
On FutureX-Past, disagreement holds a within-stratum correlation of $+0.211$ with the score, and it is
\emph{positive in all seven strata} ($+0.11$, $+0.05$, $+0.08$, $+0.36$, $+0.39$, $+0.18$, $+0.30$).
Consensus mirrors it at $-0.276$, negative in all seven. The direction is the substantive point:
\textbf{committees whose members reason differently tend to score higher, and agreement among selected
models is not in itself evidence of correctness.} This is the population-level counterpart of the medoid result---it
indicates that the gain is associated with behavioral spread in general, rather than with the particular
crowd we happened to build.

\subsection{The effect is specific to the aggregation rule}
On BTF-v3 the same signal is absent: $+0.024$ pooled and $-0.010$ within strata, with the sign varying
across bands. This is what the aggregation rules predict rather than a contradiction. BTF-v3 answers are
probabilities combined by a median, which averages disagreement away before it can act; FutureX-Past
answers are sets, and aggregation must select one of several competing candidates, which is exactly
where differing views can change the outcome. Behavioral diversity helps where the aggregation rule
gives it somewhere to act.

\subsection{An association, not yet a selection rule}
One caveat bounds the claim. The relationship above is a property of the committee population; it does
not by itself yield a recipe for choosing a committee in advance. When we hold size fixed and select by
disagreement, the chosen committee does not reliably beat a random committee of the same size---and
neither does selecting by measured score on a training split, which suggests the obstacle is that
committee-level quality transfers weakly across item samples rather than anything specific to the
label-free signal. Behavioral diversity is therefore established here as a property that accompanies
stronger crowds rather than as a criterion one can select on directly; among the constructions we tried
(Section~\ref{sec:crowd_construction}), medoid representatives are the only one that converts it into a
consistent gain.

\section{How Much of This Is the Tie-Break?}
\label{sec:tiebreak}

The FutureX-Past differences above rest on a convention that has not yet been examined, and it is
worth asking how much of them it accounts for. The two benchmarks place very different demands on
the aggregation rule.
Bench to the Future v3 elicits a probability, so a committee's answer is
the median of its members' forecasts and a tie cannot arise: the median
of a fixed multiset is a single number, and the $0.5$ threshold is a
stated rule rather than a choice. FutureX-Past is not so simple. It
elicits answer \emph{sets} at four difficulty levels, and the aggregation
that respects those types---plurality at level~1, and a medoid under set
$F_1$ or ordered overlap at levels~2--4---frequently returns several
candidates with identical support. The problem is structural rather than
incidental: a similarity medoid over exactly two candidates is tied by
construction, because each is the other's only neighbour. Across the
$K=3$ committees used here, the share of decisions with more than one
tied candidate rises with the complexity of the answer type, from
$22.3\%$ at level~1 to $55.9\%$ at level~2, $76.6\%$ at level~3 and
$80.6\%$ at level~4.

For simplicity we resolved these ties by a fixed alphabetical ordering of
the candidate answers, applied identically within a cluster (stage~1) and
between cluster decisions (stage~2). This is deterministic and
reproducible, and it is the rule behind the FutureX-Past numbers reported
elsewhere in this paper. It is also, on its face, unmotivated: the
ordering of answer strings carries no information about which answer is
correct. We therefore treat it as a hypothesis to be tested rather than a
detail to be assumed, and ask whether a better tie-break rule exists.

\subsection{Strategies and controls}

A tie-break may act at either stage of the hierarchical vote, and we
enumerate the choices available at each. At \textbf{stage~1}, when the
members of one cluster are tied: pick by a fixed ordering of the answers
(\emph{alphabetical}, and its \emph{reverse}); pick uniformly at random;
pass all tied candidates to stage~2, either splitting the cluster's vote
between them or giving each a full vote (\emph{defer}); withdraw the
cluster from stage~2 (\emph{drop cluster}); adopt the answer of the
cluster's own medoid model (\emph{cluster medoid}); or prefer the answer
held by more of the cluster's members (\emph{member count}). At
\textbf{stage~2}, when the cluster decisions are tied: the same ordering
and random rules; abstain on the question; prefer the answer backed by
the larger total cluster membership (\emph{cluster population}); or by
the greater number of clusters (\emph{cluster count}). A deployed system
must fix one policy at each stage, so we evaluate \emph{joint}
strategies---one stage-1 rule paired with one stage-2 rule---rather than
each stage in isolation.

The comparison needs a reference that is not itself a convention. We use
\textbf{random resolution at both stages}. Its expected score is exactly
the mean over the tie set,
\begin{equation}
\mathbb{E}\!\left[s(\hat{y},y)\right]
=
\frac{1}{|\mathcal{T}(q)|}\sum_{c\,\in\,\mathcal{T}(q)} s(c, y(q)),
\label{eq:tie_expectation}
\end{equation}
where $\mathcal{T}(q)$ is the set of tied candidates for question $q$, so
a rule can exceed it only by selecting better-than-average candidates.
Because the tie trees here are small enough to enumerate exhaustively, we
compute this expectation in closed form rather than by sampling, which
removes any dependence on a seed.

The null hypothesis is not that a rule has zero effect. Any deterministic
rule lands above or below the mean of Eq.~\eqref{eq:tie_expectation}
according to which candidates it happens to favour, and alphabetical
order is one such rule among many. We therefore construct the null
empirically, from $300$ \emph{arbitrary} deterministic rules obtained by
hashing each candidate answer under a different seed. These rules carry
no information about correctness by construction, and the spread of their
outcomes is the luck a convention can obtain for free. That spread is
wide: mean $-0.0038$, standard deviation $0.0262$, a 5th--95th percentile
range of $[-0.0458, +0.0398]$, and a maximum of $+0.0802$.

\subsection{Results}

Table~\ref{tab:tiebreak} reports the five strongest joint strategies
together with the fixed alphabetical ordering used elsewhere in the paper
and its reverse, each scored against random resolution on the same six
committees.

\begin{table*}[t]
\caption{Joint tie-break strategies on FutureX-Past, scored against RANDOM resolution at both stages. $\Delta$ is the mean change in the tier-weighted score across the 6 two-stage committees ($K=3$, $k\in\{2,3,4\}$, both clusterings); \emph{wins} counts the committees on which the strategy scores above random. $p$ is measured against 300 ARBITRARY deterministic rule pairs, which is the luck available to a convention that carries no information about correctness. Random is the reference and is therefore exactly zero. No strategy separates from the arbitrary-rule null.}
\label{tab:tiebreak}
\centering
\small
\begin{tabular}{llrrr}
\toprule
Stage 1 & Stage 2 & Wins & $\Delta$ vs random & $p$ \\
\midrule
defer, split weight      & reverse alphabetical   & 6/6 & $+0.0192$ & 0.191 \\
defer, full weight       & reverse alphabetical   & 6/6 & $+0.0176$ & 0.218 \\
cluster medoid           & reverse alphabetical   & 5/6 & $+0.0146$ & 0.245 \\
reverse alphabetical     & cluster population     & 6/6 & $+0.0121$ & 0.275 \\
reverse alphabetical     & alphabetical           & 5/6 & $+0.0116$ & 0.283 \\
alphabetical             & alphabetical           & 0/6 & $-0.0222$ & 0.757 \\
reverse alphabetical     & reverse alphabetical   & 4/6 & $+0.0104$ & 0.303 \\
\midrule
random (reference) & random & --- & $0.0000$ & --- \\
\bottomrule
\end{tabular}
\end{table*}

Three findings follow, and they are consistent with one another.

\textbf{The fixed ordering is not the best strategy, and is not even a
good one.} Alphabetical order at both stages is the weakest
non-abstaining strategy in the grid: it scores below random on all six
committees, with a mean of $-0.0222$. Simply reversing the same ordering
moves the strategy to $+0.0104$ and above random on four of six. A rule
whose direction can be flipped for a swing of $0.033$ is not measuring
anything about the answers; the two directions are equally defensible and
equally uninformative.

\textbf{Random resolution is the neutral reference, and the deterministic
rules straddle it.} The alphabetical rule sits below random, its reverse
above, and the arbitrary-rule null spans $[-0.0458, +0.0398]$ at the 5th
and 95th percentiles. Random is not a weak baseline to be beaten but the
centre of the distribution that arbitrary conventions are drawn from.

\textbf{No deterministic strategy separates from that null.} The best
joint strategy gains $+0.0192$, while an arbitrary pair of rules reaches
$+0.0802$; every strategy in Table~\ref{tab:tiebreak} has $p \geq 0.19$.
Consistency does not rescue them either. Three strategies score above
random on all six committees, which a coin-flip model would call
significant at $p = 0.016$---but the committees share a question set and
overlapping members, and an arbitrary rule sweeps all six
$10.0\%$ of the time. Judged against that, a clean sweep carries
$p = 0.103$. It is also telling \emph{which} strategies sweep: three of
the five strongest pair a principled rule with the \emph{reverse}
alphabetical ordering, an arbitrary rule by construction.

We conclude that tie resolution is not a productive axis for improving
LLM crowd prediction. The ties are real and frequent, and the choice
among conventions moves the reported score by more than the differences
between the crowd-construction strategies this paper compares---but no
rule we could devise extracts signal from them. Random resolution is therefore the
defensible default: not the best rule, because none is, but an unbiased
one that no reader has to accept on faith. For reporting it can be taken
in closed form as the expectation of
Eq.~\eqref{eq:tie_expectation}, which needs no seed; where a deployed
system must emit a single answer, a seeded draw gives the same guarantee
in expectation.

The FutureX-Past scores reported elsewhere in this paper were produced
with the fixed alphabetical rule, not with that expectation. This is worth
stating plainly because the measurement above puts that rule at the
\emph{unlucky} end of the range: it wins on none of the six committees
and sits $0.0222$ below random. The crowd-construction differences we
report on FutureX-Past were therefore obtained under a tie convention
that, if anything, works against them.
This finding also bounds the interpretation of the FutureX-Past results
reported above. The spread across tie conventions,
$[-0.0222, +0.0192]$ around random on the same committees, is comparable
to the differences between the crowd-construction strategies this section
compares; the BTF3 column carries no such caveat, because its median rule
admits no tie. Where the two benchmarks disagree about a strategy, the
FutureX-Past side of that disagreement is the less firmly established.
\end{document}